\documentclass[preprint,12pt]{elsarticle}

\usepackage[utf8]{inputenc}
\usepackage[english]{babel}
\usepackage{amsmath}
\usepackage{amssymb}
\usepackage{amsthm}
\usepackage{booktabs}
\usepackage{diagbox}
\usepackage{float}
\usepackage{balance}
\usepackage{textcomp}
\usepackage{stfloats}

\usepackage{subcaption}
\usepackage{framed}
\usepackage[linesnumbered,ruled,vlined,commentsnumbered]{algorithm2e}
\usepackage[normalem]{ulem}
\usepackage{arydshln}
\usepackage{mathtools}
\usepackage{amsmath}
\usepackage{amsthm}
\usepackage{amssymb}
\usepackage{amsfonts}
\usepackage{dsfont}
\usepackage{enumerate}
\usepackage{comment}
\usepackage[mathlines]{lineno}
\usepackage{csquotes}
\usepackage{bm}
\usepackage{hyperref}
\hypersetup{
    colorlinks=true,
    linkcolor=blue,
    filecolor=magenta,      
    urlcolor=cyan,
}
\newtheorem{theorem}{Theorem}[section]
\newtheorem{lemma}[theorem]{Lemma}
\newtheorem{proposition}[theorem]{Proposition}
\newtheorem{definition}[theorem]{Definition}
\newtheorem{remark}[theorem]{Remark}
\newtheorem{corollary}[theorem]{Corollary}
\newtheorem{assume}{Assumption}[section]

\newcommand{\rn}{\mathbb{R}^n}
\newcommand{\rN}{\mathbb{R}^N}
\newcommand{\Rmn}{\mathbb{R}^{m\times n}}
\newcommand{\RNn}{\mathbb{R}^{N\times n}}
\newcommand{\st}{\mathcal{S}}\newcommand{\sgn}{\mathrm{sign}}
\newcommand{\hl}{\mathcal{H}^L}
\newcommand{\opnorm}{2\rightarrow2}

\journal{Applied \& Computational Harmonic Analysis}

\begin{document}

\begin{frontmatter}



\title{Generalization analysis of an unfolding network for analysis-based Compressed Sensing}



\author[1,2]{Vicky Kouni}
\address[1]{Mathematical Institute, University of Oxford}
\ead{kouni@maths.ox.ac.uk}
\address[2]{Isaac Newton Institute for Mathematical Studies, University of Cambridge}
\ead{vk428@cam.ac.uk}

\author[3,4]{Yannis Panagakis}
\address[3]{Department of Informatics \& Telecommunications,\\ National \& Kapodistrian University of Athens}
\ead{yannisp@di.uoa.gr}
\address[4]{Archimedes AI, Athena RC}

\begin{abstract}
Unfolding networks have shown promising results in the Compressed Sensing (CS) field. Yet, the investigation of their generalization ability is still in its infancy. In this paper, we perform a generalization analysis of a state-of-the-art ADMM-based unfolding network, which jointly learns a decoder for CS and a sparsifying redundant analysis operator. To this end, we first impose a structural constraint on the learnable sparsifier, which parametrizes the network's hypothesis class. For the latter, we estimate its Rademacher complexity. With this estimate in hand, we deliver generalization error bounds -- which scale like the square root of the number of layers -- for the examined network. Finally, the validity of our theory is assessed and numerical comparisons to a state-of-the-art unfolding network are made, on synthetic and real-world datasets. Our experimental results demonstrate that our proposed framework complies with our theoretical findings and outperforms the baseline, consistently for all datasets.

\end{abstract}




\end{frontmatter}


\section{Introduction}
\label{intro}
\emph{Compressed Sensing} (CS) is a modern technique for recovering signals from incomplete, noisy observations. To date, various optimization algorithms are employed for tackling the CS problem \cite{irls}, \cite{daubechies}, \cite{fista}, \cite{boyd:5}. However, the fact that model-based methods may lack in terms of time complexity and/or reconstruction quality, has led researchers to develop data-driven approaches for dealing with CS \cite{romberg}, \cite{infocs}, \cite{transcs}. In a recent line of research, the merits of iterative methods and deep neural networks are combined in \emph{deep unfolding} \cite{unfolding}, \cite{unrolling}. The latter constitutes a technique for interpreting the iterations of optimization algorithms as layers of a neural network, which reconstructs the signals of interest from their compressive measurements.
\par Deep unfolding networks (DUNs) for inverse problems \cite{scarlett}, \cite{deadmm} are preferred to standard deep learning architectures, since they enjoy advantages like interpretability \cite{superres}, prior knowledge of signal structure \cite{wcs} and a relatively small number of trainable parameters \cite{mimo}. The same holds true in the case of CS, where state-of-the-art unfolding networks \cite{sista}, \cite{gendnn}, \cite{underwater}, \cite{unfoldenoise}, \cite{csc} typically learn a function called \emph{decoder}, which reconstructs $x$ from $y$. In fact, unfolding networks based on the \textit{iterative soft-thresholding algorithm} (ISTA \cite{daubechies}) and the \textit{alternating direction of multipliers method} (ADMM \cite{boyd:5}) seem to be the most popular classes of DUNs targeting the CS problem. Such networks can also learn -- jointly with the decoder -- a sparsifying transform for the data \cite{istagen}, \cite{ista-net}, \cite{admm-net}, \cite{ladmm}, \cite{admmdad}, \cite{tensoradmm}, integrating that way a dictionary learning technique. Due to the advantages that the latter has offered when applied in model-based methods \cite{1bitcs}, \cite{imagedl}, \cite{iot}, it seems intriguing to examine its effectiveness when combined with DUNs.
\par Nevertheless, most of the aforementioned ISTA- and ADMM-based DUNs promote synthesis sparsity \cite{rf} in their framework, since the learnable sparsifying dictionary satisfies some orthogonality constraint. Distinct from its synthesis counterpart \cite{cosparse}, the analysis sparsity model may be more advantageous for CS \cite{analvssyn}. For example, it takes into account the redundancy of the involved analysis operators, leading to a more flexible sparse representation of the signals, as opposed to orthogonal sparsifiers \cite{genzel} (see Section \ref{relwork} for a detailed comparison between the two sparsity models). To our knowledge, only one state-of-the-art  ADMM-based DUN \cite{admmdad} -- which comprises the preliminary work of this article -- solves the CS problem by entailing analysis sparsity, in terms of learning a sparsifying redundant analysis operator.
\par From the mathematical viewpoint, the generalization analysis of deep neural networks \cite{shalev}, \cite{mohri} attracts significant research interest \cite{gensgd}, \cite{wass}, \cite{arora}, \cite{spectralbound}. Nevertheless, the estimation of the generalization error of DUNs is still at an early stage. Particularly, generalization error bounds are mainly provided for the class of ISTA-based unfolding networks \cite{istagen}, \cite{deeprnn}, \cite{unfoldrnn}. To our knowledge,
the generalization ability of ADMM-based DUNs is not yet explained.
\par In this paper, distinct from the previous methods, we leverage a ``built-in'' characteristic of ADMM to impose specific structure on the learnable sparsifying redundant transform of a state-of-the-art  ADMM-based DUN, namely ADMM-DAD \cite{admmdad}. For the latter, we estimate its generalization error, in terms of the Rademacher complexity of its associated hypothesis class. In the end, we present empirical evidence supporting our derived generalization error bounds. Our contributions are summarized below.
\begin{enumerate}
    \item Inspired by recent representatives of the classes of ISTA- and ADMM-based unfolding networks \cite{istagen}, \cite{admm-net}, \cite{ladmm}, \cite{admmdad} (see Section \ref{relwork} for a brief description of a subset of them), we address the generalization analysis of a state-of-the-art  ADMM-based DUN, namely ADMM-DAD \cite{admmdad}, which deals with analysis-based CS. Towards that direction, we first exploit inherent structure of the original ADMM algorithm and impose a structural constraint on the learnable sparsifier of ADMM-DAD. Our proposed framework -- presented in Section \ref{hypoth} -- induces a frame property in the learnable redundant analysis operator, which parametrizes the hypothesis class of ADMM-DAD. To our knowledge, we are the first to impose such a structure on the hypothesis class of a DUN solving the analysis-based CS problem.
    \item In Section \ref{chain}, we employ chaining \cite[Chapter 8]{vershynin} to upper-bound the Rademacher complexity \cite{bartlett} of the hypothesis class of ADMM-DAD. Our novelty lies on studying the generalization ability of this ADMM-based DUN, by means of the afore-stated upper-bound on the Rademacher complexity. The generalization error bounds for ADMM-DAD are presented in Section \ref{genbounds}.
    \item We verify our theoretical guarantees in Section \ref{exp}, by numerically testing ADMM-DAD on a synthetic dataset and a real-world image dataset, i.e., MNIST \cite{mnist}. We also compare the performance of ADMM-DAD to that of a recent variant of ISTA-net \cite{istagen}. In all experiments, ADMM-DAD outperforms the baseline and its generalization ability conforms with our theoretical results.
\end{enumerate}
\noindent\textit{Notation.} For a sequence $(a_n)_{n\in\mathbb{N}}$ that is upper bounded by some $M>0$, we write $|a_n|\leq M$, for all $n\in\mathbb{N}$. For a matrix $A\in\mathbb{R}^{m\times n}$, we write $\|A\|_{\opnorm}$ for its spectral norm and $\|A\|_F$ for its Frobenius norm. The $l_2$-norm of a vector in $\rn$ is represented by $\|\cdot\|_2$. We write $X\in\mathbb{R}^{n\times s}$ for the matrix containing the data points $x_1,x_2,\hdots,x_s\in\rn$ as its columns; similarly, we write $Y\in\mathbb{R}^{m\times s}$ for the matrix collecting the measurements $y_1,y_2,\hdots,y_s\in\mathbb{R}^m$. For functions $f:\mathbb{R}^m\mapsto\rn$, we denote by $f(Y)$ the matrix whose $i$th column is $f(y_i)$. We call \textit{analysis operator} the linear mapping $\Phi:\rn\mapsto\rN$, with associated matrix $\Phi\in\RNn$, whose action on any vector $x\in\rn$ is given by $\Phi x:=\{\langle x,\Phi_i\rangle\}_{i=1}^N$, where $\Phi_i$, $i=1,\hdots,N$, are the rows of $\Phi$. Without loss of generality, we will interchangeably use the term ``analysis operator'' to indicate either the mapping or its associated matrix.
The adjoint of $\Phi$, i.e. $\Phi^T$, is the \textit{synthesis operator}. Moreover, the rows of $\Phi$ constitute a \textit{frame} for $\rn$ if it holds
    \begin{equation}
    \alpha\|x\|_2^2\leq\sum_{i=1}^N\lvert\langle x,\Phi_i\rangle\rvert^2\leq \beta \|x\|_2^2
\end{equation}
for all $x\in\rn$, for some $0<\alpha\leq\beta<\infty$ (frame bounds); $\alpha$ is the \emph{lower frame bound} and $\beta$ is the \textit{upper frame bound}. We denote with $S$ the multiplication of a synthesis with an analysis operator, i.e. $S=\Phi^T\Phi\in\mathbb{R}^{n\times n}$, and call it $S$-operator. Moreover, if $S$ is invertible, then the rows of $\Phi$ constitute a frame and we call $S$ the \emph{frame operator} associated with that frame. For the frame operator and its inverse, it holds $\alpha\leq\|S\|_{\opnorm}\leq \beta$ and $\beta^{-1}\leq\|S^{-1}\|_{\opnorm}\leq\alpha^{-1}$, respectively. For matrices $A_1,A_2\in\mathbb{R}^{N\times N}$, we denote by $[A_1;A_2]\in\mathbb{R}^{2N\times N}$ their concatenation with respect to the first dimension, while we denote by $[A_1\,|\,A_2]\in\mathbb{R}^{N\times 2N}$ their concatenation with respect to the second dimension. We write $O_{N\times N}$ for a real-valued $N\times N$ matrix filled with zeros and $I_{N\times N}$ for the $N\times N$ identity matrix. For $x\in\mathbb{R},\,\tau>0$, the soft thresholding operator $\st_\tau:\mathbb{R}\mapsto\mathbb{R}$ is defined as
    \begin{equation}\label{sto}
    \st_\tau(x)=\st(x,\tau)=
    \begin{cases}
    \sgn(x)(|x|-\tau),& |x|\geq\tau\\
    0,&\text{otherwise,}
    \end{cases}
    \end{equation} or in closed form $\st(x,\tau)=\sgn(x)\max(0,|x|-\tau)$. For $x\in\rn$, the soft thresholding operator acts component-wise, i.e. $(\st_\tau(x))_i=\st_\tau(x_i)$, and is 1-Lipschitz continuous with respect to $x$ under the $\ell_2$-norm. For $y\in\rn,\,\tau>0$, the mapping
    \begin{equation}\label{proxmap}
        P_G(\tau;y)=\mathrm{argmin}_{x\in\rn}\left\{\tau G(x)+\frac{1}{2}\|x-y\|_2^2\right\},
    \end{equation}
    is the \textit{proximal mapping associated to the convex function G}. For $G(\cdot)=\|\cdot\|_1$, \eqref{proxmap} coincides with \eqref{sto}. For two functions $f,g:\mathbb{R}^n\mapsto\mathbb{R}^n$, we write their composition as $f\circ g:\mathbb{R}^n\mapsto\mathbb{R}^n$ and if there exists some constant $C>0$ such that $f(x)\leq Cg(x)$, then we write $f(x)\lesssim g(x)$ for all $x\in\rn$. For the ball of radius $t>0$ in $\rn$ with respect to some norm $\|\cdot\|$, we write $B_{\|\cdot\|}^n(t)$. The covering number $\mathcal{N}(T,d,t)$ of a metric space $(T,d)$ at level $t>0$, is defined as the smallest number of balls with respect to the metric $d$ required to cover $T$. When the metric is induced by some norm $\|\cdot\|$, we write $\mathcal{N}(T,\|\cdot\|,t)$.

\section{Background on sparsity models and unfolding networks for CS}\label{relwork}
\subsection{Synthesis-based CS: unfolding ISTA and ADMM}
CS aims at recovering $x\in\rn$ from $y=Ax+e\in\mathbb{R}^m$, $m<n$, with $A$ being the measurement matrix and $e\in\mathbb{R}^m$, $\|e\|_2\leq\epsilon$, corresponding to noise. To do so, one can impose a synthesis sparsity model on $x$ \cite{rf}, \cite{waveletcs}, i.e., assume that there exists $D\in\mathbb{R}^{n\times p}$ ($n\leq p$) such that $x=Dz$, with the coefficients' vector $z\in\mathbb{R}^p$ being sparse. In fact, $D$ is typically chosen to be an orthogonal matrix, e.g. a wavelet or cosine transform. By incorporating synthesis sparsity in CS, one is called to solve the LASSO problem:
\begin{equation}\label{lasso}
    \min_{z\in\mathbb{R}^p}\frac{1}{2}\|y-\Tilde{A}z\|_2^2+\lambda\|z\|_1,
\end{equation}
with $\Tilde{A}=AD$ and $\lambda>0$ being a regularization parameter. Two broad classes of algorithms that are commonly employed to solve \eqref{lasso} rely on ISTA and ADMM. These methods incorporate a proximal mapping \eqref{proxmap} and yield iterative schemes which, under mild assumptions, output a minimizer $\hat{z}$ of \eqref{lasso}; then the desired reconstructed signal is simply given by $\hat{x}=D\hat{z}$. If $D$ is regarded unknown and learned from a sequence of training samples, the iterations of ISTA and ADMM are interpreted as layers of neural networks; such DUNs are usually coined ISTA-nets \cite{istagen}, \cite{lista} and ADMM-nets \cite{admm-net, robustadmmnet}, respectively. They jointly learn sparsifying transforms for the data and a decoder for CS, that is, a function reconstructing $x$ from $y$.

\subsection{Analysis-based CS: unfolding ADMM}\label{analcs}

The algorithms and corresponding DUNs we described so far rely on the synthesis sparsity model, since their framework incorporates some orthogonality constraint for the learnable sparsifiers. A tractable counterpart of synthesis sparsity is the analysis sparsity model \cite{robustanal}, \cite{star}, \cite{genzel}, (also known as cosparse model \cite{cosparse}, \cite{cosparsity}). In the latter, one assumes that there exists some \textit{redundant analysis operator} $\Phi\in\RNn$, $N>n$, such that $\Phi x$ is sparse. Under the analysis sparsity model, the optimization problem for CS is formulated as a generalized LASSO problem, i.e.,
\begin{equation}
    \label{genlasso}
    \min_{x\in\rn}\frac{1}{2}\|Ax-y\|_2^2+\lambda\|\Phi x\|_1.
\end{equation}
Particularly, analysis sparsity has gained research interest, due to some advantages it may offer compared to its synthesis counterpart. For example, the redundancy of an analysis operator associated to a frame \cite{christensen} can provide greater -- than orthonormal bases -- flexibility in the sparse representation of signals \cite{casazza}. Moreover, it is computationally more efficient to use sparsifying redundant transforms instead of orthogonal ones, since the iterative algorithm for CS may need less measurements $m$ for perfect reconstruction \cite{genzel}. Last but not least, in the case of synthesis sparsity with $n<p$ for $D\in\mathbb{R}^{n\times p}$, one can argue that it is preferable to solve \eqref{genlasso}, since the dimension of the optimization problem is smaller \cite{analvssyn}. Now, thresholding algorithms like ISTA cannot treat analysis sparsity, since the proximal mapping associated to $\|\Phi(\cdot)\|_1$ does not have a closed-form solution. Therefore, we turn to ADMM, which can efficiently solve \eqref{genlasso} by means of the following iterative scheme:
\begin{align}
\label{xup}
    &x^{k+1}=(A^TA+\rho \Phi^T\Phi)^{-1}(A^Ty+\rho \Phi^T(z^k-u^k))\\
    \label{zup}
        &z^{k+1}=\st_{\lambda/\rho}(\Phi x^{k+1}-u^k)\\
        \label{uup}
        &u^{k+1}=u^k+\Phi x^{k+1}-z^{k+1},
\end{align}

with $x^k\in\rn$ and $z^k,u^k\in\mathbb{R}^N$, for all $k\in\mathbb{N}$, and scalar $\rho>0$ being the so-called penalty parameter, satisfying the role of a step-size for the corresponding updates \cite{boyd:5}. Following a standard setup for ADMM \cite{boyd:5}, we consider all of the vectors' initial values, i.e., $x^0$, $z^0$ and $u^0$, to be the zero vector of corresponding dimension.\\
To unroll ADMM into a network, we begin with substituting \eqref{xup} into the update rules \eqref{zup} and \eqref{uup}, and then \eqref{zup} into \eqref{uup}. After that, we concatenate the resulting variables $z^k,u^k$ in one \textit{intermediate variable}, i.e., $v_k=[u^k;z^k]\in\mathbb{R}^{2N}$, so that we arrive at
\begin{equation}\label{v2}
    v_{k+1}=\Tilde{\Theta}v_k+I'b+I''\st_{\lambda/\rho}(\Theta v_k+b),
\end{equation}
where
\begin{align}
    \Theta&=[-I-W\,|\,W]\in\mathbb{R}^{N\times 2N}\\
    W&=\rho\Phi R^{-1}\Phi^T\in\mathbb{R}^{N\times N}\\
    R&=A^TA+\rho\Phi^T\Phi\in\mathbb{R}^{n\times n}\\
    \Tilde{\Theta}&=[\Lambda;O_{N\times 2N}]\in\mathbb{R}^{2N\times 2N}\\
    \Lambda&=[I-W\,|\,W]\in\mathbb{R}^{N\times 2N}\\
    I'&=[I_{N\times N};O_{N\times N}]\in\mathbb{R}^{2N\times N}\\
    I''&=[-I_{N\times N};I_{N\times N}]\in\mathbb{R}^{2N\times N}
\end{align}
and
\begin{align}
    b=b(y)&=\Phi\tau(y)\in\mathbb{R}^{N}\\
    \label{toft}
    \tau=\tau(y)&=R^{-1}A^Ty\in\rn
\end{align}
For more details on the unrolling procedure, we refer the interested reader to \cite{admmdad}.\\
To enable a learning scenario, we assume that the redundant analysis operator $\Phi$ is unknown and learned from a set of i.i.d. training samples, i.e. $\mathbf{S}=\{(x_i,y_i)\}_{i=1}^s$, drawn from an unknown distribution\footnote{Formally speaking, this is a distribution over $x_i$ and for fixed $A,e$, we obtain $y_i=Ax_i+e$} $\mathcal{D}^s$. Then, the updates in \eqref{xup} - \eqref{uup} can be interpreted as a neural network with $L\in\mathbb{N}$ layers, coined ADMM Deep Analysis Decoding (ADMM-DAD) \cite{admmdad}. Based on \eqref{v2}, the output of the first and the $k$th layer are given\footnote{Although \eqref{v2} is equivalent to the functional form of the outputs given in \eqref{layer1} and \eqref{layerk}, we prefer the latter since they will facilitate our generalization analysis presented in Section~\ref{mainsec}} by
\begin{align}
    \label{layer1}
    f_1(y) & =I'b+I''\st_{\lambda/\rho}(b),\\
    \label{layerk}
    f_k(v) & =\Tilde{\Theta}v+I'b+I''\st_{\lambda/\rho}(\Theta v+b), \quad k = 2, \hdots, L,
\end{align}
The composition of $L$ such layers (all having the same $\Phi$) is denoted by
\begin{equation}\label{interdec}
    f^L_{\Phi}(y)=f_{L}\circ\dots\circ f_1(y)
\end{equation}
and constitutes an \textit{intermediate decoder} -- realized by ADMM-DAD -- that reconstructs $v$ from $y$. Motivated by \eqref{xup}, we acquire the desired solution $\hat{x}$ by applying an affine map $T_\Phi:\mathbb{R}^{2N}\mapsto\mathbb{R}^{n}$ after the final layer $L$, so that
\begin{equation}\label{tauf}
    \hat{x}=T_\Phi(v):=C_\Phi v+\tau,
\end{equation}
where
\begin{align}
\label{coft}
    C_\Phi&=[-\rho R^{-1}\Phi^T\,\vert\,\rho R^{-1}\Phi^T]\in\mathbb{R}^{n\times2N}.
\end{align}
Finally, the application of an appropriate clipping function
\begin{equation}\label{psi}
    \sigma(x) = \left\{\begin{array}{cc}
    x,&\ \|x\|_2 \leq B_{\mathrm{out}}\\
    B_{\mathrm{out}}\frac{x}{\|x\|_2},&\ \text{otherwise}
    \end{array}\right.,
\end{equation} 
for some fixed constant $B_{\mathrm{out}}>0$, so that the output is pushed inside a reasonable range of values, yields the desired decoder, i.e.,
\begin{equation}\label{decoder}
    \mathrm{dec}_\Phi^L(y)=\sigma(T_\Phi(f_\Phi^L(y))),
\end{equation}
implemented by ADMM-DAD.

\section{Generalization Analysis of ADMM-DAD}
\label{mainsec}
\subsection{Enhancing the hypothesis class of ADMM-DAD}\label{hypoth}

According to \cite{admmdad}, the hypothesis class of ADMM-DAD consists of all the decoders that ADMM-DAD can implement and is parametrized by the learnable redundant analysis operator $\Phi$:
\begin{equation}
    \label{hypo}
    \mathcal{H}^L=\{ h:\,\mathbb{R}^m\mapsto\mathbb{R}^n:h(y)=\sigma(T_\Phi(f^L_{\Phi}(y))),\,\Phi\in\mathbb{R}^{N\times n},N>n\}.
\end{equation}
However, the definition of \eqref{hypo} does not account for any particular structure on $\Phi$, which in turn could explain the performance of ADMM-DAD. On the other hand, the $x$-update \eqref{xup} of ADMM incorporates the term $S=\Phi^T\Phi$, which is typically assumed to be an invertible matrix \cite{boyd:5}, \cite{genlasso}. Similarly, typical choices for the measurement matrix $A$ consist of a (appropriately normalized) Gaussian matrix \cite{rf}, \cite{kr}, \cite{cosparse}. Therefore, we are inspired by the aforementioned facts and make some assumptions, which will hold for the rest of the paper.
\begin{assume}\label{assume1}
    For an analysis operator $\Phi\in\RNn$ with $N>n$, the matrix $S=\Phi^T\Phi$ is invertible.
\end{assume}
\begin{assume}\label{assume2}
    For an analysis operator satisfying Assumption \ref{assume1}, and for appropriately chosen measurement matrix $A\in\Rmn$ and penalty parameter $\rho>0$, it holds $\rho\|S^{-1}\|_{\opnorm}\|A\|_{\opnorm}<1$.
\end{assume}

\begin{remark}
    From a theoretical perspective, it is reasonable to incorporate the invertibility of $S$ in our framework, since the set of non-invertible matrices $S$ of the form $S=\Phi^T\Phi$ has zero Lebesgue measure.
    Additionally, Assumptions \ref{assume1} and \ref{assume2} are empirically confirmed (see Section \ref{expinv}), since ADMM-DAD learns a $\Phi$ with associated $S$-operator satisfying $S^{-1}S=I$ and $\rho\|S^{-1}\|_{\opnorm}\|A\|_{\opnorm}<1$.
\end{remark}

Due to Assumption \ref{assume1}, we can further assume that there exists some $0<\beta<\infty$, so that $\|S\|_{\opnorm}\leq\beta$, which leads us to introduce the following definition.

\begin{definition}\label{frame}
    We define $\mathcal{F}_{\beta }$ to be the class of redundant analysis operators $\Phi\in\RNn$ for which the associated $S$-operator is invertible and has bounded spectral norm by some $0<\beta<\infty$.
\end{definition}

\begin{remark}\label{phibound}
    The invertibility of $S$ in Definition \ref{frame} implies that the rows of $\Phi$ constitute a frame for $\rn$. Hence, $S$ is a frame operator and for some $0<\alpha\leq\beta<\infty$, it holds $\alpha\leq\|S\|_{\opnorm}\leq\beta$ and $\|\Phi\|_{\opnorm}\leq\sqrt{\beta}$.
\end{remark}
We restrict the hypothesis class $\hl$ of ADMM-DAD to be parameterized by $\Phi\in\mathcal{F}_{\beta}$, in order to account for a structural constraint on $\Phi$.

\begin{definition}
    We define $\mathbf{H}^L\subset\hl$ to be the hypothesis class of ADMM-DAD, i.e., the space of all the decoders that ADMM-DAD can implement, parameterized by $\Phi\in\mathcal{F}_{\beta }$:
\begin{equation}\label{hypothesis}
    \mathbf{H}^L=\{ h:\mathbb{R}^m\mapsto\rn:\, h(y)=\sigma(T_\Phi(f^L_{\Phi}(y))),\,\Phi\in\mathcal{F}_{\beta }\}.
\end{equation}
\end{definition} 
Given the hypothesis class \eqref{hypothesis} and the training set $\mathbf{S}$, ADMM-DAD yields $h\in\mathbf{H}^L$ such that $h(y)=\hat{x}\approx x$. For a loss function $\ell:\mathbf{H}^L\times\rn\times\mathbb{R}^m\mapsto\mathbb{R}_{>0}$, we define the empirical loss of a hypothesis $h\in\mathbf{H}^L$ as
\begin{equation}\label{empiricalrisk}
    \mathcal{\hat{L}}_{train}(h)=\frac{1}{s}\sum_{i=1}^s\ell(h,x_i,y_i).
\end{equation}
For the rest of the paper, we work with $\ell(\cdot)=\|\cdot\|_2^2$, so that \eqref{empiricalrisk} transforms into the \emph{training mean-squared error} (train MSE):
\begin{equation}\label{trainmse}
    \mathcal{\hat{L}}_{train}(h)=\frac{1}{s}\sum_{j=1}^s\|h(y_j)-x_j)\|_2^2.
\end{equation}
We also define the \textit{true loss} to be
\begin{equation}\label{truerisk}
    \mathcal{L}(h)=\mathbb{E}_{(x,y)\sim\mathcal{D}}(\|h(y)-x\|_2^2).
\end{equation}
The difference between \eqref{trainmse} and \eqref{truerisk}, i.e.,
\begin{equation}\label{generror}
    \text{GE}(h)=|\mathcal{\hat{L}}_{train}(h)-\mathcal{L}(h)|,
\end{equation}
constitutes the \emph{generalization error} of ADMM-DAD and informs us about how well the network performs on unseen data. Since $\mathcal{D}$ is unknown, we estimate \eqref{generror} in terms of the \emph{empirical Rademacher complexity} \cite{shalev}:
\begin{equation}
    \label{erc}
    \mathcal{R}_{\mathbf{S}}(\ell\circ\mathbf{H}^L)=\mathbb{E}\sup_{h\in\mathbf{H}^L}\frac{1}{s}\sum_{i=1}^s\epsilon_i\|h(y_i)-x_i\|_2^2,
\end{equation}
with $\epsilon$ being a Rademacher vector, i.e, a vector with i.i.d. entries taking the values $\pm1$ with equal probability. To do so, we deploy the next Theorem.

\begin{theorem}[{\cite[Theorem 26.5]{shalev}}]\label{radem}
Let $\mathcal{H}$ be a family of functions, $\mathcal{S}$ the training set drawn from $\mathcal{D}^s$, and $\ell$ a real-valued bounded loss function satisfying $|\ell(h,z)|\leq c$, for all $h\in\mathcal{H}, z\in Z$. Then, for $\delta\in(0,1)$, with probability at least $1-\delta$, we have for all $h\in\mathcal{H}$
\begin{equation}
    \mathcal{L}(h)\leq\mathcal{\hat{L}}_{train}(h)+2\mathcal{R}_{\mathcal{S}}(\ell\circ\mathcal{H})+4c\sqrt{\frac{2\log(4\delta)}{s}}.
\end{equation}
\end{theorem}

In order to apply the latter in $\mathbf{H}^L$, we prove that $\|\cdot\|_2^2$ is bounded by some constant $c>0$. Towards that direction, we make two typical -- for the machine learning literature -- assumptions for $\mathbf{S}$. Let us suppose that with overwhelming probability it holds:
\begin{equation}
    \label{ybound}
    \|y_i\|_2\leq\mathrm{B_{in}},
\end{equation}
for some constant $\mathrm{B_{in}}>0$, $i=1,2,\hdots,s$. We also assume that for any $h\in\mathbf{H}^L$, with overwhelming probability over $y_i$ chosen from $\mathcal{D}$, it holds
\begin{equation}\label{hbound}
    \|h(y_i)\|_2\leq B_{\text{out}},
\end{equation}
by definition of $\sigma$, for some constant $B_{\text{out}}>0$, $i=1,2,\hdots,s$. Then, we have $\|h(y_i)-x_i\|_2^2\leq (B_{\text{in}}+B_{\text{out}})^2$, for all $i=1,2,\hdots,s$; hence, $c=(B_{\text{in}}+B_{\text{out}})^2$.\\
We simplify the quantity $\mathcal{R}_{\mathbf{S}}(\|\cdot\|_2^2\circ\mathbf{H}^L)$, by using the following contraction principle for vector-valued functions\footnote{Lemma~\ref{contraction} constitutes a simpler and more concise version of a more general result given in \cite{tal}}:
\begin{lemma}[{\cite[Corollary 4]{contraction}}]
\label{contraction}
Let $\mathcal{H}$ be a set of functions $h:\mathcal{X}\mapsto\mathbb{R}^n$, $f:\mathbb{R}^n\mapsto\mathbb{R}^n$ a $K$-Lipschitz function and $\mathcal{S}=\{x_i\}_{i=1}^s$. Then
\begin{equation}
    \mathbb{E}\sup_{h\in\mathcal{H}}\sum_{i=1}^s\epsilon_if\circ h(x_i)\leq\sqrt{2}K\mathbb{E}\sup_{h\in\mathcal{H}}\sum_{i=1}^s\sum_{k=1}^n\epsilon_{ik}h_k(x_i),
\end{equation}
where $(\epsilon_i)$ and $(\epsilon_{ik})$ are Rademacher sequences.
\end{lemma}
Lemma~\ref{contraction} allows us to study $\mathcal{R}_\mathbf{S}(\mathbf{H}^L)$ alone. Since it is easy to check that $\|\cdot\|^2_2$ is Lipschitz continuous, with Lipschitz constant $\mathrm{Lip}_{\|\cdot\|^2_2}=2\mathrm{B_{in}}+2\mathrm{B_{out}}$, we employ Lemma \ref{contraction} to obtain:
\begin{align}
    \mathcal{R}_\mathbf{S}(l\circ\mathbf{H}^L)&\leq\sqrt{2}(2\mathrm{B_{in}}+2\mathrm{B_{out}})\mathbb{E}\sup_{h\in\mathbf{H}^L}\sum_{i=1}^s\sum_{k=1}^n\epsilon_{ik}h_k(y_i)\notag\\
    \label{contradem}
    &=\sqrt{2}(2\mathrm{B_{in}}+2\mathrm{B_{out}})\mathcal{R}_\mathbf{S}(\mathbf{H}^L).
\end{align}
Therefore, we are left with estimating \eqref{contradem}. We do so in a series of steps, presented in the next subsections, with the main parts of our proof strategy inspired by \cite{istagen}.

\subsection{Bounded outputs}\label{boundsec}

We pass to matrix notation by accounting for the number of samples in the training set $\mathbf{S}$. Hence, we apply the Cauchy-Schwartz inequality in \eqref{ybound}, \eqref{hbound} yielding
\begin{align}
\label{boundedin}
\|Y\|_F&\leq\sqrt{s}\mathrm{B}_{\text{in}},\\
\label{boundedout}
\|h(Y)\|_F&=\|\sigma(T_\Phi(f_{\Phi}^L(Y)))\|_F\leq\sqrt{s}\mathrm{B}_{\text{out}},
\end{align}
respectively. We also state below two results that will be needed in some of the proofs later on.

\begin{lemma}[Proof in the supplementary material]\label{invab}
    Let $A\in\mathbb{R}^{n\times n}$ be invertible and $B\in\mathbb{R}^{n\times n}$. For a sub-multiplicative matrix norm $\|\cdot\|$ on $\mathbb{R}^{n\times n}$, if it holds $\|A^{-1}\|\|B\|<1$, then $A+B\in\mathbb{R}^{n\times n}$ is invertible. Moreover, we have
    \begin{equation}
        \|(A+B)^{-1}\|\leq\frac{\|A^{-1}\|}{1-\|A^{-1}\|\|B\|}.
    \end{equation}
\end{lemma}

\begin{lemma}[Proof in the supplementary material]\label{invsubtract}
    For a sub-multiplicative matrix norm $\|\cdot\|$ on $\mathbb{R}^{n\times n}$, if $A,\,B\in\mathbb{R}^{n\times n}$ are invertible, then
    \begin{equation}
        \|B^{-1}-A^{-1}\|\leq\|B^{-1}\|\|A^{-1}\|\|A-B\|.
    \end{equation}
\end{lemma}

We prove that the output of the intermediate decoder \eqref{interdec} is bounded with respect to the Frobenius norm, after any number of layers $k<L$.

\begin{proposition}\label{boundedoutput}
    Let $k\in\mathbb{N}$. For any $\Phi\in\mathcal{F}_{\beta }$, with $\mathcal{F}_{\beta }$ as in Definition~\ref{frame}, and arbitrary $\lambda,\,\rho>0$ in the definition of $f_\Phi^k$, we have
    \begin{equation}\label{fbound}
        \|f_\Phi^k(Y)\|_F\leq3\|A\|_{\opnorm}\|Y\|_Fq\sqrt{\beta }\sum_{i=0}^{k-1}3^i(1+2q\rho\beta)^i,
    \end{equation}
where $q=\frac{\rho}{\alpha-\rho\|A^TA\|_{\opnorm}}$, and $\alpha$ and $\beta$ are defined as in Remark~\ref{phibound}.
\end{proposition}

\begin{proof}
    We prove \eqref{fbound} via induction. For $k=1$:
    \begin{equation}\label{f1}
        \|f_\Phi^1(Y)\|_F\leq3\|B\|_F\leq3\|A\|_{\opnorm}\|Y\|_F\sqrt{\beta }\|(A^TA+\rho\Phi^T\Phi)^{-1}\|_{\opnorm},
    \end{equation}
which holds by definition of \eqref{layer1}. The invertibility of $S=\Phi^T\Phi$, along with Assumption \ref{assume2}, Remark \ref{phibound} and Lemma \ref{invab}, imply that
\begin{align}
    \|(A^TA+\rho\Phi^T\Phi)^{-1}\|_{\opnorm}&=\|(A^TA+\rho S)^{-1}\|_{\opnorm}\leq\frac{\rho\|S^{-1}\|_{\opnorm}}{1-\rho\|S^{-1}\|_{\opnorm}\|A^TA\|_{\opnorm}}\notag\\ 
    \label{invs}
    &=\frac{\rho}{\alpha-\rho\|A^TA\|_{\opnorm}}:=q,
\end{align}
where in the last step we used the fact that $\beta^{-1}\leq\|S^{-1}\|_{\opnorm}\leq\alpha^{-1}$, for some $0<\alpha\leq\beta<\infty$. Substituting \eqref{invs} into \eqref{f1} yields $\|f_\Phi^1(Y)\|_F\leq3\|A\|_{\opnorm}\|Y\|_Fq\sqrt{\beta}$. Suppose now that \eqref{fbound} holds for some $k\in\mathbb{N}$. Then, for $k+1$:
\begin{align*}
    \|f_\Phi^{k+1}(Y)\|_F\leq&\|\Tilde{\Theta}\|_{\opnorm}\|f_\Phi^k(Y)\|_F+2\|\Theta\|_{\opnorm}\|f_\Phi^k(Y)\|_F+3\|B\|_F\\
    \leq&3\left((1+2\|W\|_{\opnorm})\|f_\Phi^k(Y)\|_F+\|B\|_F\right)\\
    \leq&3\Bigg((1+2q\rho\beta )\left(3\|A\|_{\opnorm}\|Y\|_Fq\sqrt{\beta}\sum_{i=0}^{k-1}3^i(1+2q\rho\beta )^i\right)\\
    &+\|A\|_{\opnorm}\|Y\|_Fq\sqrt{\beta}\Bigg)\\
    =&3\|A\|_{\opnorm}\|Y\|_Fq\sqrt{\beta}\sum_{i=0}^{k}3^i(1+2q\rho\beta )^i.
\end{align*}
The proof follows.
\end{proof}

\subsection{Lipschitzness with respect to $\Phi$}\label{lipsec}
With the previous result in hand, we prove that the intermediate decoder \eqref{interdec} and the final decoder \eqref{decoder} are Lipschitz continuous with respect to $\Phi$.

\begin{theorem}[Proof in the supplemental material]
\label{lipschitz}
    Let $f^L_W$ defined as in \eqref{interdec}, $L\geq2$, and analysis operator $\Phi\in\mathcal{F}_{\beta}$, with $\mathcal{F}_{\beta }$ as in Definition~\ref{frame}. Then, for any $\Phi_1,\,\Phi_2\in\mathcal{F}_{\beta }$, it holds
    \begin{equation}
    \|f^L_{\Phi_1}(Y)-f^L_{\Phi_2}(Y)\|_F\leq K_L\|\Phi_1-\Phi_2\|_{2\rightarrow2},
\end{equation} where
\begin{multline}
\label{fpert}
    K_L=qG^L+\sum_{k=2}^L\Bigg(G^{L-k}\bigg[qG+36\beta q^2\rho(1+\beta q\rho)\|A\|_{\opnorm}\|Y\|_F\sum_{i=0}^{k-2}G^i\bigg]\Bigg),
\end{multline}
with $G=3(1+2\beta q\rho)$, $q$ as in Proposition~\ref{boundedoutput}, and $\beta$ as in Remark~\ref{phibound}.
\end{theorem}

\begin{corollary}\label{lipdec}
    Let $h\in\mathbf{H}^L$ defined as in \eqref{hypothesis}, $L\geq2$, and analysis operator $\Phi\in\mathcal{F}_{\beta }$, with $\mathcal{F}_{\beta }$ as in Definition~\ref{frame}. Then, for any $\Phi_1,\,\Phi_2\in\mathcal{F}_{\beta }$, we have:
    \begin{equation}
    \|\sigma(T_{\Phi_1}(f^L_{\Phi_1}(Y)))-\sigma(T_{\Phi_2}(f^L_{\Phi_2}(Y)))\|_F\leq\Sigma_L\|\Phi_2-\Phi_1\|_{\opnorm},
\end{equation}
where
\begin{equation}\label{sigmal}
    \Sigma_L=2q\rho\sqrt{\beta}\left(K_L+3\|A\|_{\opnorm}\|Y\|_Fq(1+2\beta q\rho)\sum_{k=0}^{L-1}3^k(1+2\beta q\rho)^k\right),
\end{equation}
with $K_L$ as in Theorem~\ref{lipschitz}, $q$ as in Proposition~\ref{boundedoutput}, and $\beta$ as in Remark~\ref{phibound}.

\begin{proof}
    By definition, $\sigma$ is a 1-Lipschitz function. The affine maps $T_{\Phi_1}$ and $T_{\Phi_2}$ are also Lipschitz continuous, with Lipschitz constants $\mathrm{Lip}_{T_{\Phi_1}}$ and $\mathrm{Lip}_{T_{\Phi_2}}$, respectively, satisfying
    \begin{align}
        \label{lipT1}
        \mathrm{Lip}_{T_{\Phi_1}}=\mathrm{Lip}_{T_{\Phi_2}}=\|T_{\Phi_1}\|_{\opnorm}=\|T_{\Phi_2}\|_{\opnorm}\leq2q\rho\sqrt{\beta },
    \end{align}
since $\Phi_1,\,\Phi_2\in\mathcal{F}_{\beta }$, and due to the explicit forms of \eqref{toft} and \eqref{coft}, with $q$ as in Proposition~\ref{boundedoutput} and $\beta$ as in Remark~\ref{phibound}. Applying Proposition~\ref{boundedoutput} and \eqref{lipT1}, as well as Theorem \ref{lipschitz}, in the last step of the following derivation, yields
\begin{align*}
    \|\sigma(&T_{\Phi_1}(f^L_{\Phi_1}(Y)))-\sigma(T_{\Phi_2}(f^L_{\Phi_2}(Y)))\|_F\\
    \leq&\|T_{\Phi_1}(f^L_{\Phi_1}(Y))-T_{\Phi_2}(f^L_{\Phi_2}(Y))\|_F\\
    =&\|T_{\Phi_1}(f^L_{\Phi_1}(Y))-T_{\Phi_1}(f^L_{\Phi_2}(Y))+T_{\Phi_1}(f^L_{\Phi_2}(Y))-T_{\Phi_2}(f^L_{\Phi_2}(Y))\|_F\\
    \leq&\|T_{\Phi_1}\|_{\opnorm}\|f^L_{\Phi_2}(Y))-f^L_{\Phi_1}(Y))\|_F+\|T_{\Phi_2}-T_{\Phi_1}\|_{\opnorm}\|f^L_{\Phi_1}(Y))\|_F\\
    \leq&2q\rho\sqrt{\beta }K_L\|\Phi_2-\Phi_1\|_{\opnorm}+\left(3\|A\|_{\opnorm}\|Y\|_Fq\sqrt{\beta}\sum_{k=0}^{L-1}G^k\right)\|T_{\Phi_2}-T_{\Phi_1}\|_{\opnorm},
\end{align*}
where $G=3(1+2q\rho\beta)$. The introduction of mixed terms and the application of Lemma \ref{invsubtract} give:
\begin{align*}
    \|T_{\Phi_2}&-T_{\Phi_1}\|_{\opnorm}\\
    \leq&2\rho\|(A^TA+\rho\Phi_2^T\Phi_2)^{-1}\Phi^T_{\Phi_2}-(A^TA+\rho\Phi_1^T\Phi_1)^{-1}\Phi^T_{\Phi_1}\|_{\opnorm}\\
    =&2\rho\|(A^TA+\rho\Phi_2^T\Phi_2)^{-1}\Phi^T_{\Phi_2}-(A^TA+\rho\Phi_2^T\Phi_2)^{-1}\Phi^T_{\Phi_1}\\
    &+(A^TA+\rho\Phi_2^T\Phi_2)^{-1}\Phi^T_{\Phi_1}-(A^TA+\rho\Phi_1^T\Phi_1)^{-1}\Phi^T_{\Phi_1}\|_{\opnorm}\\
    \leq&2\rho\bigg(q\|\Phi_2-\Phi_1\|_{\opnorm}+\sqrt{\beta }\|(A^TA+\rho\Phi_2^T\Phi_2)^{-1}-(A^TA+\rho\Phi_1^T\Phi_1)^{-1}\|_{\opnorm}\bigg)\\
    \leq&2\rho\bigg(q\|\Phi_2-\Phi_1\|_{\opnorm}+2\beta q^2\rho\|\Phi_2-\Phi_1\|_{\opnorm}\bigg)\\
    =&2q\rho(1+2q\beta\rho)\|\Phi_2-\Phi_1\|_{\opnorm}.
\end{align*}
Overall, we obtain
\begin{equation}
    \|\sigma(T_{\Phi_1}(f^L_{\Phi_1}(Y)))-\sigma(T_{\Phi_2}(f^L_{\Phi_2}(Y)))\|_F\leq\Sigma_L\|\Phi_2-\Phi_1\|_{\opnorm},
\end{equation}
where $\Sigma_L=2q\rho\sqrt{\beta}\left(K_L+\|A\|_{\opnorm}\|Y\|_FqG\sum_{k=0}^{L-1}G^k\right)$.
\end{proof}

\end{corollary}
\subsection{Chaining the Rademacher complexity}\label{chain}

We apply the results of Sections \ref{boundsec} and \ref{lipsec} and estimate the covering numbers of the set
\begin{align}\label{mset}
    \mathbf{M}:&=\{(h(y_1)|h(y_2)|\dots|h(y_s))\in\mathbb{R}^{n\times s}:\ h\in \mathbf{H}^{L}\}\notag\\
    &=\{\sigma(T_\Phi((f^L_\Phi(Y)))\in\mathbb{R}^{n\times s}:\ \Phi\in\mathcal{F}_{\beta }\},
\end{align}
which corresponds to the hypothesis class $\mathbf{H}^L$ defined in \eqref{hypothesis}. The columns of each $M\in\mathbf{M}$ constitute the reconstructions produced by $h\in\mathbf{H}^L$ when applied to each $y_i$, $i=1,2,\hdots,s$. Since both $\mathbf{M}$ and $\mathbf{H}^L$ are parameterized by $\Phi$, we rewrite \eqref{contradem} as follows:

\begin{equation}\label{muset}
    \mathcal{R}_\mathbf{S}(\mathbf{H}^L)=\mathbb{E}\sup_{h\in\mathbf{H}^L}\sum_{i=1}^s\sum_{k=1}^n\epsilon_{ik}h_k(y_i)=\mathbb{E}\sup_{M\in\mathbf{M}}\frac{1}{s}\sum_{i=1}^{s}\sum_{k=1}^n\epsilon_{ik}M_{ik}.
\end{equation}

The latter has subgaussian increments, so we employ Dudley's inequality \cite[Theorem 8.23]{rf} to upper bound it in terms of the covering numbers of $\mathbf{M}$. A key quantity appearing in Dudley's inequality is the radius of $\mathbf{M}$, that is,
\begin{align}
    \Delta(\mathbf{M})&=\sup_{h\in\mathbf{H}^L}\sqrt{\mathbb{E}\left(\sum_{i=1}^{s}\sum_{k=1}^n\epsilon_{ik}h_k(y_i)\right)^2}\leq\sup_{h\in\mathbf{H}^L}\sqrt{\mathbb{E}\sum_{i=1}^{s}\sum_{k=1}^n\epsilon_{ik}(h_k(y_i))^2}\notag\\
    \label{radius}
    &\leq\sup_{h\in\mathbf{H}^L}\sqrt{\sum_{i=1}^{s}\|h(y_i)\|_2^2}\overset{\eqref{boundedout}}{\leq}\sqrt{s}B_{\mathrm{out}}.
\end{align}
We combine \eqref{contradem}, \eqref{muset}, \eqref{radius} and apply Dudley's inequality to obtain

\begin{equation}\label{dudleys}
\begin{split}
    \mathcal{R}_\mathbf{S}(l\circ\mathbf{H}^L)\leq\frac{16(\mathrm{B_{in}}+\mathrm{B_{out}})}{s}\int_0^{\frac{\sqrt{s}B_{\mathrm{out}}}{2}}\sqrt{\log\mathcal{N}(\mathbf{M},\|\cdot\|_F,\varepsilon)}d\varepsilon.
\end{split}
\end{equation}
Finally, we upper-bound the quantity $\mathcal{N}(\mathbf{M},\|\cdot\|_F,\varepsilon)$.

\begin{lemma}\label{cover}
For $0<t<\infty$, the covering numbers of the ball $B_{\|\cdot\|_{\opnorm}}^{N\times n}(t)=\{X\in\mathbb{R}^{N\times n}:\,\|X\|_{\opnorm}\leq t\}$ satisfy the following for any $\varepsilon>0$:
\begin{equation}
    \mathcal{N}(B_{\|\cdot\|_{\opnorm}}^{N\times n}(t),\|\cdot\|_{\opnorm},\varepsilon)\leq\left(1+\frac{2t}{\varepsilon}\right)^{Nn}.
\end{equation}
\end{lemma}

\begin{proof}
For $|\cdot|$ denoting the volume in $\mathbb{R}^{N\times n}$, we adapt a well-known result \cite[Proposition 4.2.12]{vershynin}, in order to connect covering numbers and $|\cdot|$:
\begin{align*}
    \mathcal{N}(B_{\|\cdot\|_{\opnorm}}^{N\times n}(t),\|\cdot\|_{\opnorm},\varepsilon)&\leq\frac{|B_{\|\cdot\|_{\opnorm}}^{N\times n}(t)+(\frac{\varepsilon}{2})B_{\|\cdot\|_{\opnorm}}^{N\times n}(1)|}{|(\frac{\varepsilon}{2})B_{\|\cdot\|_{\opnorm}}^{N\times n}(1)|}=\frac{|(t+\frac{\varepsilon}{2})B_{\|\cdot\|_{\opnorm}}^{N\times n}(1)|}{|(\frac{\varepsilon}{2})B_{\|\cdot\|_{\opnorm}}^{N\times n}(1)|}\\
    &\leq\left(1+\frac{2t}{\varepsilon}\right)^{Nn}.\qedhere
\end{align*}
\end{proof}

\begin{proposition}\label{mcover}
For the covering numbers of $\mathbf{M}$ given in \eqref{mset} it holds:
\begin{equation}
    \mathcal{N}(\mathbf{M},\|\cdot\|_F,\varepsilon)\leq\left(1+\frac{2\sqrt{\beta }\Sigma_L}{\varepsilon}\right)^{Nn}.
\end{equation}
\end{proposition}

\begin{proof}
    By Definition \ref{frame} and Remark \ref{phibound} we have $\mathcal{F}_{\beta}\subset B_{\|\cdot\|_{\opnorm}}^{N\times n}(\sqrt{\beta})$. Then, the application of Lemma \ref{cover} implies for $\mathcal{F}_{\beta }$ that
    \begin{align}
        \mathcal{N}(\mathcal{F}_{\beta },\|\cdot\|_{\opnorm},\varepsilon)\leq\left(1+\frac{2\sqrt{\beta }}{\varepsilon}\right)^{Nn}.
    \end{align}
Therefore, the covering numbers of $\mathbf{M}$ are bounded as follows:
\begin{align}
    \mathcal{N}(\mathbf{M},\|\cdot\|_F,\varepsilon)&\leq\mathcal{N}(\Sigma_L\mathcal{F}_{\beta },\|\cdot\|_{\opnorm},\varepsilon)=\mathcal{N}(\mathcal{F}_{\beta },\|\cdot\|_{\opnorm},\varepsilon/\Sigma_L)\notag\\
    &\leq\left(1+\frac{2\sqrt{\beta }\Sigma_L}{\varepsilon}\right)^{Nn},
\end{align}
which is the desired estimate.
\end{proof}

\subsection{Generalization error bounds}\label{genbounds}
We combine the results of Section \ref{chain} with Theorem \ref{radem}, to deliver generalization error bounds for ADMM-DAD.
\begin{theorem}\label{gentheorem}
Let $\mathbf{H}^L$ be the hypothesis class defined in \eqref{hypothesis}. With probability at least $1-\delta$, for all $h\in\mathbf{H}^L$, the generalization error is bounded as
\begin{equation}
    \begin{split}
    \mathcal{L}(h)\leq\mathcal{\hat{L}}_{train}(h)&+8(B_{\mathrm{in}}+B_{\mathrm{out}})B_{\mathrm{out}}\sqrt{\frac{Nn}{s}}\sqrt{\log\left(e\left(1+\frac{2\sqrt{\beta }\Sigma_L}{\sqrt{s}B_{\mathrm{out}}}\right)\right)}\\   &+4(B_{\mathrm{in}}+B_{\mathrm{out}})^2\sqrt{\frac{2\log(4/\delta)}{s}},
    \end{split}
\end{equation}
with $\Sigma_L$ defined as in Corollary \ref{lipdec}.
\end{theorem}

\begin{proof}
We apply Proposition \ref{mcover} in \eqref{dudleys} to get

\begin{align}
    \mathcal{R}_\mathbf{S}(l\circ\mathbf{H}^L)&\leq\frac{16(\mathrm{B_{in}}+\mathrm{B_{out}})}{s}\int_0^{\frac{\sqrt{s}B_{\mathrm{out}}}{2}}\sqrt{\log\mathcal{N}(\mathbf{M},\|\cdot\|_F,\varepsilon)}d\varepsilon\notag\\
    &\leq\frac{16(\mathrm{B_{in}}+\mathrm{B_{out}})}{s}\int_0^{\frac{\sqrt{s}B_{\mathrm{out}}}{2}}\sqrt{Nn\log\left(1+\frac{2\sqrt{\beta }\Sigma_L}{\varepsilon}\right)}d\varepsilon\notag\\
    \label{radest}
    &\leq8(\mathrm{B_{in}}+\mathrm{B_{out}})B_{\mathrm{out}}\sqrt{\frac{Nn}{s}}\sqrt{\log\left(e\left(1+\frac{4\sqrt{\beta }\Sigma_L}{\sqrt{s}B_{\mathrm{out}}}\right)\right)},
\end{align}
where in the last step we used the following inequality\footnote{The interested reader may refer to \cite[Lemma C.9]{rf} for a detailed proof of this inequality}:
\begin{equation*}
    \int_0^a\sqrt{\log\left(1+\frac{b}{t}\right)}dt\leq a\sqrt{\log(e(1+b/a))},\qquad a,b>0.
\end{equation*}
We substitute the upper-bound \eqref{radest} in Theorem \ref{radem} and the proof follows.
\end{proof}

\begin{theorem}\label{gengentheorem}
Let $\mathbf{H}^L$ be the hypothesis class defined in \eqref{hypothesis}. Assume there exist pair-samples $\{(x_i,y_i)\}_{i=1}^s\overset{\text{i.i.d.}}{\sim}\mathcal{D}^s$, with $y_i=Ax_i+e$, $\|e\|_2\leq\varepsilon$, for some $\varepsilon>0$. Let us further assume that it holds $\|y_i\|_2\leq\mathrm{B_{in}}$ almost surely with $\mathrm{B_{in}}=\mathrm{B_{out}}$ in \eqref{psi}. Then with probability at least $1-\delta$, for all $h\in\mathbf{H}^L$, the generalization error is bounded as
\begin{equation}
    \begin{split}\label{genbound}
    \mathcal{L}(h)\leq\mathcal{\hat{L}}_{train}(h)+16B^2_{\mathrm{out}}\Bigg(\sqrt{\frac{Nn}{s}}&\sqrt{\log\left(e\left(1+\frac{2\sqrt{\beta }\Sigma_L}{\sqrt{s}B_{\mathrm{out}}}\right)\right)}\\
    &+\sqrt{\frac{2\log(4/\delta)}{s}}\Bigg),
    \end{split}
\end{equation}
with $\Sigma_L$ defined as in Corollary \ref{lipdec}.
\end{theorem}
\begin{remark}
    Notice that $L$ enters at most exponentially in the definition of $K_L$ \eqref{fpert} -- and thus $\Sigma_L$ \eqref{sigmal}. If we treat all terms in \eqref{genbound} as constants, except for $L$, $N$, $s$, then the previous Theorem tells us that the generalization error of ADMM-DAD roughly scales like $\sqrt{NL/s}$.
\end{remark}
\noindent\textbf{Comparison with related work}: Similarly to Theorem~\ref{gengentheorem}, the analysis depicted in \cite[Theorem 2]{istagen} demonstrates that the generalization error of the proposed synthesis-sparsity-based ISTA-net roughly scales like $\sqrt{(n\log L)(n+m)/s}$. The latter upper-bound is slightly better in terms of $L$ than our theoretical results. On the other hand, Theorem~\ref{gengentheorem} showcases that our bound does not depend on $m$, which makes it tighter in terms of the number of measurements. What is more, the generalization error\footnote{The work presented in \cite{deconet} was conducted in parallel with the present paper, and published after the present paper's submission} of a state-of-the-art -- albeit architecturally different -- analysis-sparsity-based DUN \cite{deconet} roughly scales like $\sqrt{L}$. This fact highlights the state-of-the-art generalization performance of our proposed ADMM-DAD. 

\section{Experiments}
\label{exp}
We train and test ADMM-DAD on a synthetic dataset of random vectors, drawn from the normal distribution (70000 training and 10000 test examples) and the MNIST dataset \cite{mnist}, containing 60000 training and 10000 test $28\times28$ image examples. For the MNIST dataset, we take the vectorized form of the images. We examine ADMM-DAD for alternating number of layers $L$ and redundancy ratios $N/n$. For the measurement process, we select an appropriately normalized Gaussian matrix $A\in\Rmn$, with $m/n=25\%$ CS ratio. We also add zero-mean Gaussian noise $e$, with standard deviation std $=10^{-4}$ to the measurements, so that $y=Ax+e$. We perform (He) normal initialization \cite{he} for $W\in\RNn$. We implement all models in PyTorch \cite{pytorch} and train them using the \emph{Adam} algorithm \cite{adam}, with batch size $128$. For all experiments, we report the \emph{test MSE}:
\begin{equation}\label{testmse}
    \mathcal{L}_{test}=\frac{1}{d}\sum_{i=1}^d\|h(\tilde{y}_i)-\tilde{x}_i\|_2^2,
\end{equation}
where $\mathbf{D}=\{(\tilde{y}_i,\tilde{x}_i)\}_{i=1}^d$ 
is a set of $d$ test data, that are not used during training, and the \emph{empirical generalization error} (EGE)
\begin{equation}\label{genmse}
    \mathcal{L}_{gen}=|\mathcal{L}_{test}-\mathcal{L}_{train}|,
\end{equation}
where $\mathcal{L}_{train}$ is defined in \eqref{trainmse}. Since \eqref{testmse} approximates the true loss, we use \eqref{genmse} -- which can be explicitly computed -- to approximate \eqref{generror}. We train all models, on all datasets, employing early stopping \cite{earlystop} with respect to \eqref{genmse}. We repeat all the experiments at least 10 times and average the results over the runs. We also compare ADMM-DAD to a recent variant of ISTA-net \cite{istagen}. Both DUNs learn corresponding decoders for CS, but ISTA-net promotes synthesis sparsity, by learning an orthogonal sparsifying transform; ADMM-DAD, in constrast, promotes analysis sparsity by means of the learnable redundant analysis operator. Therefore, the structure of ISTA-net makes it a nice candidate for comparison with ADMM-DAD, in order to showcase how the reconstructive and generalization ability of DUNs are affected, when employing a redundant sparsifier instead of an orthogonal one. For ISTA-net, we set the best hyper-parameters proposed by the original authors.
\begin{figure*}
    \centering
    \begin{subfigure}{1.0\textwidth}
    \includegraphics[width=\textwidth]{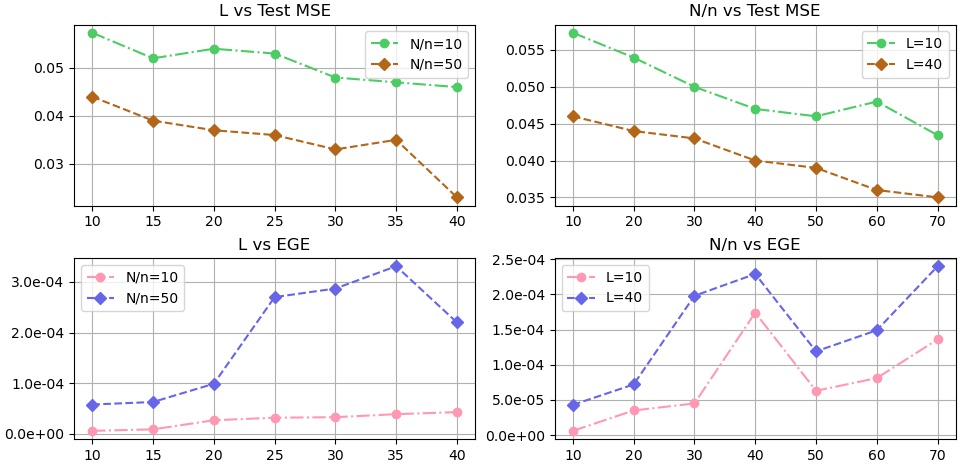}
    \captionsetup{justification=centering}
     \caption{Test MSEs and EGEs for MNIST}
    \label{mnist_plots}
    \end{subfigure}
    
    \medskip
    
    \begin{subfigure}{1.0\textwidth}
        \centering
        \includegraphics[width=\textwidth]{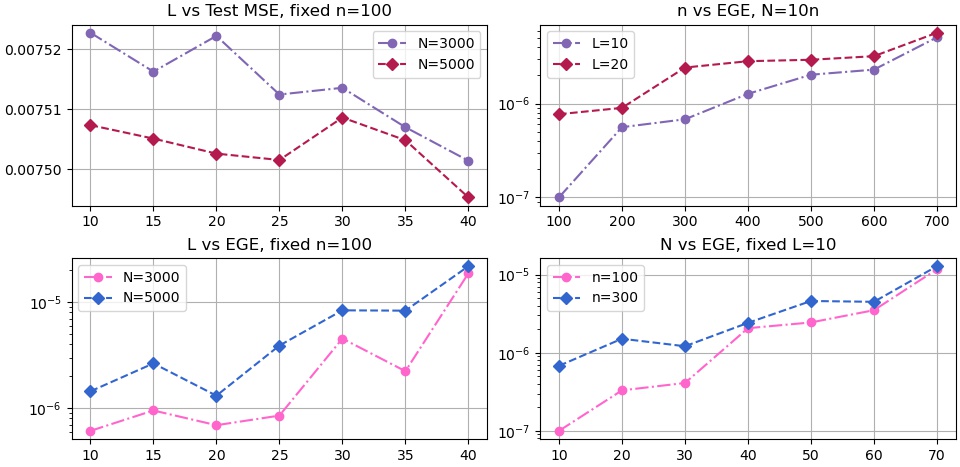}
        \captionsetup{justification=centering}
     \caption{Test MSEs and EGEs for synthetic data}
    \label{synthetic_plots}
    \end{subfigure}
    \caption{Performance plots of ADMM-DAD on (a) MNIST and (b) synthetic datasets, for varying $L$, $N$ (and $n$).}
    \label{synth}
\end{figure*}

\subsection{Experimental results \& discussion}
We evaluate the quality of our theoretical results with the following experimental scenarios.

\subsubsection{Varying $N$, $L$ on real-world image data}\label{realexp}
We examine the performance of ADMM-DAD on MNIST dataset, with varying number of layers $L$ and redundancy $N$ of the learnable sparsifier. We gather the results in Figure~\ref{mnist_plots}, which illustrates that the test MSE achieved by each instance of ADMM-DAD drops, as $L$ and $N$ increase. The decays seem reasonable, if examined from a model-based point of view. Specifically, when an iterative algorithm solves the generalized LASSO problem \eqref{genlasso}, it is expected that the reconstruction quality and performance of the solver will benefit from the (high) redundancy offered by the involved analysis operators \cite{genzel}, especially as the number of iterations/layers increases. On the other hand, the EGE of ADMM-DAD increases as both $L$ and $N/n$ increase. This behaviour confirms the theory we developed in Section \ref{genbounds}, since the EGE seems to scale like $\sqrt{NL}$.

\begin{table*}[h!]
    \centering
    \scalebox{0.7}{\begin{tabular}{||c|c|c|c|c|c|c|c|c|c||}\hline
    & \multicolumn{8}{|c|}{Test MSE}\\
         \hline
         Dataset & \multicolumn{3}{|c|}{Synthetic} & \multicolumn{3}{|c|}{MNIST} \\
         \hline
         \diagbox{Decoder}{Layers} & $L=10$ & $L=20$ & $L=30$ & $L=10$ & $L=20$ & $L=30$ \\ \hline
         ADMM-DAD (Ours) & \bf 0.007725 & \bf 0.007600 & \bf 0.007586 & \bf 0.046391 & \bf 0.040282 & \bf 0.032001 \\
         \hline
         ISTA-net \cite{istagen} & 0.007959 & 0.007774 & 0.007710 & 0.070645 & 0.068006 & 0.066325\\
         \hline
    \end{tabular}}
    
    \medskip
    \scalebox{0.7}{\begin{tabular}{||c|c|c|c|c|c|c|c|c|c||}\hline
    &  \multicolumn{8}{|c|}{Generalization Error}\\
         \hline
         Dataset & \multicolumn{3}{|c|}{Synthetic} & \multicolumn{3}{|c|}{MNIST} \\
         \hline
         \diagbox{Decoder}{Layers} & $L=10$ & $L=20$ & $L=30$ & $L=10$ & $L=20$ & $L=30$ \\ \hline
         ADMM-DAD (Ours) & \bf $\mathbf{0.22\cdot10^{-6}}$ & $\mathbf{1.04\cdot10^{-6}}$ & $\mathbf{1.65\cdot10^{-6}}$ & $\mathbf{0.63\cdot10^{-4}}$ & $\mathbf{0.40\cdot10^{-4}}$ & $\mathbf{1.21\cdot10^{-4}}$ \\
         \hline
         ISTA-net \cite{istagen} & $4.48\cdot10^{-6}$ & $2.64\cdot10^{-6}$ & $9.44\cdot10^{-6}$ & $22.51\cdot10^{-4}$ & $50.45\cdot10^{-4}$ & $76.16\cdot10^{-4}$\\
         \hline
    \end{tabular}}
    \caption{Test MSEs and empirical generalization errors for 10-, 20- and 30-layer decoders, with fixed $25\%$ CS ratio and redundancy ratio $N/n=50$. Bold letters indicate the best performance between the two decoders.}
    \label{decoders}
\end{table*}

\subsubsection{Varying $n$, $N$, $L$ on synthetic data}\label{synexp}
We test ADMM-DAD on a synthetic dataset, with varying $L$, $N$ and ambient dimension $n$. We report the results in Figure~\ref{synthetic_plots}, which illustrates the reconstruction error decreasing as $L$ increases. Regarding the generalization error, we observe in Figure~\ref{synthetic_plots} that the EGE appears to grow at the rate of $\sqrt{nNL}$, despite the fact that the theoretical generalization error bounds depend on other terms as well. The overall performance of ADMM-DAD again conforms with our theoretical results.

\subsubsection{Comparison to baseline}\label{baselinecomp}
We examine how analysis and synthesis sparsity models affect the generalization ability of unfolding networks solving the CS problem. To that end, we compare the decoders of ADMM-DAD and ISTA-net, on the MNIST and the synthetic datasets, for varying number of layers. For the synthetic dataset, we fix the ambient dimension to $n=300$. For ADMM-DAD, we set $N=39200$ for the sparsifier acting on the MNIST dataset and $N=15000$ for the sparsifier acting on the synthetic data. Our results are collected in Table~\ref{decoders}. As depicted in the latter, ADMM-DAD's decoder outperforms ISTA-net's decoder, consistently for both datasets, in terms of both reconstruction and generalization error. For the former, our experiments confirm the model-based results regarding the advantage of analysis sparsity over its synthesis counterpart (cf. Section \ref{analcs}). As for the generalization error: our results indicate that the redundancy of the learnable sparsifier acts beneficially for the generalization ability of ADMM-DAD, compared to the orthogonality of ISTA-net's framework.

\subsubsection{A note on the invertibility of $S=\Phi^T\Phi$}\label{expinv}
\begin{figure*}
    \centering
    \begin{subfigure}{1.0\textwidth}
    \includegraphics[width=\textwidth]{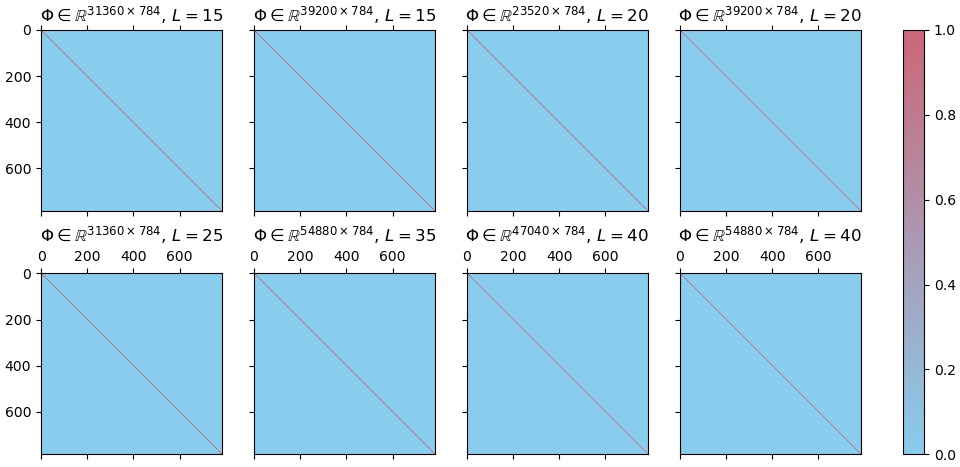}
    \captionsetup{justification=centering}
     \caption{MNIST dataset}
    \label{vis_mnist}
    \end{subfigure}
    
    \medskip
    
    \begin{subfigure}{1.0\textwidth}
        \centering
        \includegraphics[width=\textwidth]{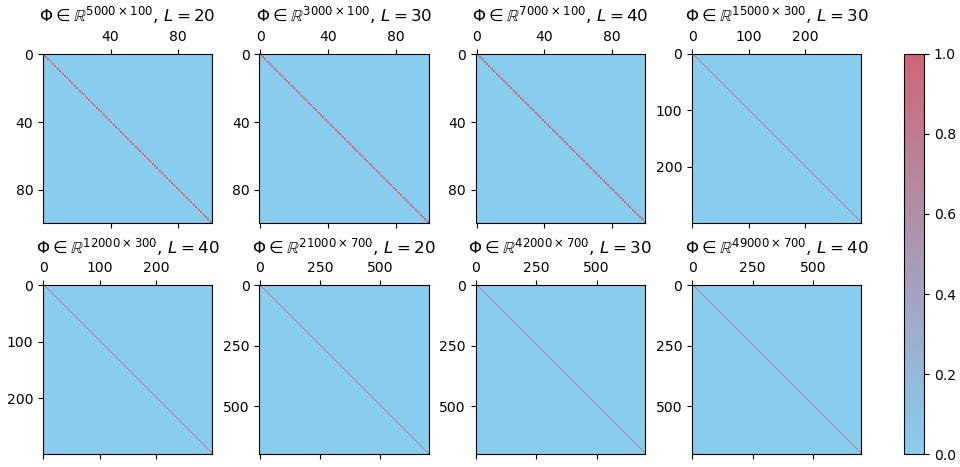}
        \captionsetup{justification=centering}
     \caption{Synthetic data}
    \label{vis_syn}
    \end{subfigure}
    \caption{Visualization of $S^{-1}S$ on (a) MNIST and (b) synthetic datasets, for varying $L$, $N$ (and $n$) of the associated learnable $\Phi$.}
    \label{visualizations}
\end{figure*}

We revisit the setups of Sections \ref{realexp}, \ref{synexp} and implement exemplary instances of ADMM-DAD, in order to verify Assumptions \ref{assume1} and \ref{assume2}. To that end, we examine the values of $S^{-1}S$ and  $\rho\|S^{-1}\|_{\opnorm}\|A\|_{\opnorm}$, for fixed $\rho=0.1$, $\|A\|_{\opnorm}\approx2$ and with $S$-operator associated to each learned $\Phi$, and present the results in Figure \ref{visualizations} and Table~\ref{svalues}, respectively. According to the latter, the values of $\rho\|S^{-1}\|_{\opnorm}\|A\|_{\opnorm}$ are consistently less than 1, for different tuples of $L$, $N$ (and $n$), which is in accordance to Assumption \ref{assume2}. We also provide in Figure \ref{visualizations} a visualization of the structure of $S^{-1}S$. As illustrated in the aforementioned figure, ADMM-DAD learns a redundant analysis operator $\Phi$ with associated $S$-operator satisfying\footnote{Due to Python's round-off errors, we consider the identity matrix $I$ to have ones on the main diagonal and non-diagonal entries of the order at most $10^{-5}$} $S^{-1}S=I$. This observation validates our intuition for imposing Assumption \ref{assume1} in our framework, as well as constraining $\Phi$ to lie in $\mathcal{F}_\beta$ (see Section \ref{hypoth}). Furthermore, we conjecture that the fact that ADMM-DAD learns an analysis operator associated to a frame could explain its increased performance, compared to the synthesis-based baseline; this assumption could serve as a potential line of future work. Note that we have also conducted experiments with a regularizer of the form $\|S^{-1}S - I\|_F$, in order to cover the small probability of learning a $\Phi$ such that $S$ is not invertible. Since ADMM-DAD with and without the regularizer yielded almost identical performance, we chose to proceed with minimizing the train MSE only. Overall, this set of example experiments showcases that the appearance of the term $\Phi^T\Phi$ in the iterative scheme of ADMM-DAD, induces a frame property to the learnable redundant analysis operator $\Phi$.

\begin{table*}
    \centering
    \scalebox{0.8}{\begin{tabular}{||c|c||}
    \hline
    $(N,\,L)$ & $\rho\|S^{-1}\|_{\opnorm}\|A\|_{\opnorm}$\\
    \hline\hline
    (23520, 10) & 0.0003\\ 
    \hline
    (31360, 15) & 0.0002\\
    \hline
    (39200, 15) & 0.0002\\
    \hline
    (23520, 20) & 0.0003\\
    \hline
    (39200, 20) & 0.0002\\ 
    \hline
    (31360, 25) & 0.0002\\
    \hline
    (7840, 30) & 0.0021\\
    \hline
    (15680, 30) & 0.0009\\
    \hline
    (54880, 35) & 0.0001\\
    \hline
    (47040, 40) & 0.0001\\
    \hline
    (54880, 40) & 0.0001\\
    \hline
    \end{tabular}}
    \medskip
    \scalebox{0.8}{\begin{tabular}{||c|c||}
    \hline
    $(n,N,L)$ & $\rho\|S^{-1}\|_{\opnorm}\|A\|_{\opnorm}$\\
    \hline\hline
    (100, 1000, 10) & 0.0031\\
    \hline
    (100, 5000, 20) & 0.0004\\
    \hline
    (100, 3000, 30) & 0.0007\\
    \hline
    (100, 7000, 40) & 0.0002\\
    \hline
    (300, 21000, 10) & 0.0003\\
    \hline
    (300, 6000, 20) & 0.0007\\
    \hline
    (300, 15000, 30) & 0.0002\\
    \hline
    (300, 12000, 40) & 0.0003\\
    \hline
    (700, 21000, 20) & 0.0002\\
    \hline
    (700, 42000, 30) & 0.0001\\
    \hline
    (700, 49000, 40) & 0.00008\\
    \hline
    \end{tabular}}
    \caption{Examination of the values of $\rho\|S^{-1}\|_{\opnorm}\|A\|_{\opnorm}$, under different choices of $L$, $N$ (and $n$), for the MNIST (left) and the synthetic (right) datasets.}
    \label{svalues}
\end{table*}

\section{Conclusion and Future Work}

In this paper, we studied the generalization ability of a state-of-the-art ADMM-based unfolding network, namely ADMM-DAD. The latter jointly learns a decoder for Compressed Sensing (CS) and a sparsifying redundant analysis operator. To that end, we first exploited an inherent characteristic of ADMM to impose a meaningful structural constraint on ADMM-DAD's learnable sparsifier; the latter parametrized ADMM-DAD's hypothesis class. Our novelty relies on the fact that the proposed framework induces a frame property on the learnable sparsifying transform. Then, we employed chaining to estimate the Rademacher complexity of ADMM-DAD's hypothesis class. With this estimate in hand, we delivered generalization error bounds for ADMM-DAD. To our knowledge, we are the first to study the generalization ability of an ADMM-based unfolding network, that solves the analysis-based CS problem. Finally, we conducted experiments validating our theory and compared ADMM-DAD to a state-of-the-art unfolding network for CS; the former outperformed the latter, consistently for all datasets. As a future line of work, we would like to include mores experiments regarding the structure of ADMM-DAD, especially with respect to the afore-stated frame property. Additionally, it would be interesting to include numerical comparisons among ADMM-DAD and ADMM-based unfolding networks promoting synthesis sparsity in CS.

\section*{Acknowledgements}
V. Kouni would like to thank the Isaac Newton Institute for Mathematical Sciences for supporting her during her INI Postdoctoral Research Fellowship in the Mathematical Sciences, especially during the programme ``Representing, calibrating \& leveraging prediction uncertainty from statistics to machine learning''. This work was funded by the EPSRC (Grant Number EP/V521929/1).\\
A part of this research was conducted during V. Kouni's affiliation with the Dep. of Informatics \& Telecommunications. During that period, V. Kouni acknowledged financial support for the implementation of this paper by Greece and the European Union (European Social Fund-ESF) through the Operational Program ``Human Resources Development, Education and Lifelong Learning'' in the context of the Act ``Enhancing Human Resources Research Potential by undertaking a Doctoral Research'' Sub-action 2: IKY Scholarship Program for PhD candidates in the Greek Universities.

\section*{Conflict of Interest Statement}
On behalf of all authors, the corresponding author states that there is no conflict of interest.







\bibliographystyle{elsarticle_num_names}
\bibliography{ref}

\begin{thebibliography}{61}
\expandafter\ifx\csname natexlab\endcsname\relax\def\natexlab#1{#1}\fi
\providecommand{\url}[1]{\texttt{#1}}
\providecommand{\href}[2]{#2}
\providecommand{\path}[1]{#1}
\providecommand{\DOIprefix}{doi:}
\providecommand{\ArXivprefix}{arXiv:}
\providecommand{\URLprefix}{URL: }
\providecommand{\Pubmedprefix}{pmid:}
\providecommand{\doi}[1]{\href{http://dx.doi.org/#1}{\path{#1}}}
\providecommand{\Pubmed}[1]{\href{pmid:#1}{\path{#1}}}
\providecommand{\bibinfo}[2]{#2}
\ifx\xfnm\relax \def\xfnm[#1]{\unskip,\space#1}\fi
\bibitem[{Chartrand and Yin(2008)}]{irls}
\bibinfo{author}{R.~Chartrand}, \bibinfo{author}{W.~Yin},
\newblock \bibinfo{title}{Iteratively reweighted algorithms for compressive
  sensing},
\newblock in: \bibinfo{booktitle}{Int. Conf. Acoust., Speech and Signal
  Process.}, \bibinfo{organization}{IEEE}, \bibinfo{year}{2008}, pp.
  \bibinfo{pages}{3869--3872}.
\bibitem[{Daubechies et~al.(2004)Daubechies, Defrise, and De~Mol}]{daubechies}
\bibinfo{author}{I.~Daubechies}, \bibinfo{author}{M.~Defrise},
  \bibinfo{author}{C.~De~Mol},
\newblock \bibinfo{title}{An iterative thresholding algorithm for linear
  inverse problems with a sparsity constraint},
\newblock \bibinfo{journal}{Commun. Pure and Appl. Math.} \bibinfo{volume}{57}
  (\bibinfo{year}{2004}) \bibinfo{pages}{1413--1457}.
\bibitem[{Beck and Teboulle(2009)}]{fista}
\bibinfo{author}{A.~Beck}, \bibinfo{author}{M.~Teboulle},
\newblock \bibinfo{title}{A fast iterative shrinkage-thresholding algorithm for
  linear inverse problems},
\newblock \bibinfo{journal}{SIAM J. Imag. Sci.} \bibinfo{volume}{2}
  (\bibinfo{year}{2009}) \bibinfo{pages}{183--202}.
\bibitem[{Boyd et~al.(2011)Boyd, Parikh, and Chu}]{boyd:5}
\bibinfo{author}{S.~Boyd}, \bibinfo{author}{N.~Parikh},
  \bibinfo{author}{E.~Chu}, \bibinfo{title}{Distributed optimization and
  statistical learning via the alternating direction method of multipliers},
  \bibinfo{publisher}{Now Publishers Inc}, \bibinfo{year}{2011}.
\bibitem[{Xu et~al.(2019)Xu, Zeng, and Romberg}]{romberg}
\bibinfo{author}{S.~Xu}, \bibinfo{author}{S.~Zeng},
  \bibinfo{author}{J.~Romberg},
\newblock \bibinfo{title}{Fast compressive sensing recovery using generative
  models with structured latent variables},
\newblock in: \bibinfo{booktitle}{Int. Conf. Acoust., Speech and Sig.
  Process.}, \bibinfo{organization}{IEEE}, \bibinfo{year}{2019}, pp.
  \bibinfo{pages}{2967--2971}.
\bibitem[{Liu and Scarlett(2020)}]{infocs}
\bibinfo{author}{Z.~Liu}, \bibinfo{author}{J.~Scarlett},
\newblock \bibinfo{title}{Information-theoretic lower bounds for compressive
  sensing with generative models},
\newblock \bibinfo{journal}{IEEE J. Selected Areas Inf. Theory}
  \bibinfo{volume}{1} (\bibinfo{year}{2020}) \bibinfo{pages}{292--303}.
\bibitem[{Shen et~al.(2022)Shen, Gan, Ning, Hua, and Zhang}]{transcs}
\bibinfo{author}{M.~Shen}, \bibinfo{author}{H.~Gan}, \bibinfo{author}{C.~Ning},
  \bibinfo{author}{Y.~Hua}, \bibinfo{author}{T.~Zhang},
\newblock \bibinfo{title}{Transcs: A transformer-based hybrid architecture for
  image compressed sensing},
\newblock \bibinfo{journal}{IEEE Trans. Image Process.} \bibinfo{volume}{31}
  (\bibinfo{year}{2022}) \bibinfo{pages}{6991--7005}.
\bibitem[{Hershey et~al.(2014)Hershey, Roux, and Weninger}]{unfolding}
\bibinfo{author}{J.~R. Hershey}, \bibinfo{author}{J.~L. Roux},
  \bibinfo{author}{F.~Weninger},
\newblock \bibinfo{title}{Deep unfolding: Model-based inspiration of novel deep
  architectures},
\newblock \bibinfo{journal}{arXiv preprint arXiv:1409.2574}
  (\bibinfo{year}{2014}).
\bibitem[{Monga et~al.(2021)Monga, Li, and Eldar}]{unrolling}
\bibinfo{author}{V.~Monga}, \bibinfo{author}{Y.~Li}, \bibinfo{author}{Y.~C.
  Eldar},
\newblock \bibinfo{title}{Algorithm unrolling: Interpretable, efficient deep
  learning for signal and image processing},
\newblock \bibinfo{journal}{IEEE Signal Process. Mag.} \bibinfo{volume}{38}
  (\bibinfo{year}{2021}) \bibinfo{pages}{18--44}.
\bibitem[{Scarlett et~al.(2022)Scarlett, Heckel, Rodrigues, Hand, and
  Eldar}]{scarlett}
\bibinfo{author}{J.~Scarlett}, \bibinfo{author}{R.~Heckel},
  \bibinfo{author}{M.~R. Rodrigues}, \bibinfo{author}{P.~Hand},
  \bibinfo{author}{Y.~C. Eldar},
\newblock \bibinfo{title}{Theoretical perspectives on deep learning methods in
  inverse problems},
\newblock \bibinfo{journal}{IEEE J. Sel. Areas in Inf. Theory}
  \bibinfo{volume}{3} (\bibinfo{year}{2022}) \bibinfo{pages}{433--453}.
\bibitem[{An et~al.(2022)An, Yue, Liu, Shang, and Liu}]{deadmm}
\bibinfo{author}{W.~An}, \bibinfo{author}{Y.~Yue}, \bibinfo{author}{Y.~Liu},
  \bibinfo{author}{F.~Shang}, \bibinfo{author}{H.~Liu},
\newblock \bibinfo{title}{A numerical {DE}s perspective on unfolded linearized
  admm networks for inverse problems},
\newblock in: \bibinfo{booktitle}{Proc. of the 30th {ACM} Int. Conf.
  Multimedia}, \bibinfo{year}{2022}, pp. \bibinfo{pages}{5065--5073}.
\bibitem[{Zhou et~al.(2023)Zhou, Yan, Pan, Ren, Xie, and Cao}]{superres}
\bibinfo{author}{M.~Zhou}, \bibinfo{author}{K.~Yan}, \bibinfo{author}{J.~Pan},
  \bibinfo{author}{W.~Ren}, \bibinfo{author}{Q.~Xie}, \bibinfo{author}{X.~Cao},
\newblock \bibinfo{title}{Memory-augmented deep unfolding network for guided
  image super-resolution},
\newblock \bibinfo{journal}{Int. J. Comput. Vis.} \bibinfo{volume}{131}
  (\bibinfo{year}{2023}) \bibinfo{pages}{215--242}.
\bibitem[{Zhang et~al.(2020)Zhang, Li, Yu, Gu, Cheng, and Gong}]{wcs}
\bibinfo{author}{J.~Zhang}, \bibinfo{author}{Y.~Li}, \bibinfo{author}{Z.~L.
  Yu}, \bibinfo{author}{Z.~Gu}, \bibinfo{author}{Y.~Cheng},
  \bibinfo{author}{H.~Gong},
\newblock \bibinfo{title}{Deep unfolding with weighted $l_2$ minimization for
  compressive sensing},
\newblock \bibinfo{journal}{IEEE Internet of Things J.} \bibinfo{volume}{8}
  (\bibinfo{year}{2020}) \bibinfo{pages}{3027--3041}.
\bibitem[{Hu et~al.(2020)Hu, Cai, Shi, Xu, Yu, and Ding}]{mimo}
\bibinfo{author}{Q.~Hu}, \bibinfo{author}{Y.~Cai}, \bibinfo{author}{Q.~Shi},
  \bibinfo{author}{K.~Xu}, \bibinfo{author}{G.~Yu}, \bibinfo{author}{Z.~Ding},
\newblock \bibinfo{title}{Iterative algorithm induced deep-unfolding neural
  networks: Precoding design for multiuser mimo systems},
\newblock \bibinfo{journal}{IEEE Trans. Wirel. Commun.} \bibinfo{volume}{20}
  (\bibinfo{year}{2020}) \bibinfo{pages}{1394--1410}.
\bibitem[{Wisdom et~al.(2017)Wisdom, Powers, Pitton, and Atlas}]{sista}
\bibinfo{author}{S.~Wisdom}, \bibinfo{author}{T.~Powers},
  \bibinfo{author}{J.~Pitton}, \bibinfo{author}{L.~Atlas},
\newblock \bibinfo{title}{Building recurrent networks by unfolding iterative
  thresholding for sequential sparse recovery},
\newblock in: \bibinfo{booktitle}{Int. Conf. Acoust., Speech and Signal
  Process.}, \bibinfo{organization}{IEEE}, \bibinfo{year}{2017}, pp.
  \bibinfo{pages}{4346--4350}.
\bibitem[{Mou et~al.(2022)Mou, Wang, and Zhang}]{gendnn}
\bibinfo{author}{C.~Mou}, \bibinfo{author}{Q.~Wang},
  \bibinfo{author}{J.~Zhang},
\newblock \bibinfo{title}{Deep generalized unfolding networks for image
  restoration},
\newblock in: \bibinfo{booktitle}{Proc. Conf. Comput. Vision and Pattern
  Recogn.}, \bibinfo{organization}{IEEE}, \bibinfo{year}{2022}, pp.
  \bibinfo{pages}{17399--17410}.
\bibitem[{Yang et~al.(2022)Yang, Xiao, and Deligiannis}]{underwater}
\bibinfo{author}{Y.~Yang}, \bibinfo{author}{P.~Xiao},
  \bibinfo{author}{N.~Deligiannis},
\newblock \bibinfo{title}{Underwater localization with binary measurements:
  From compressed sensing to deep unfolding},
\newblock \bibinfo{journal}{Digital Signal Process.}  (\bibinfo{year}{2022})
  \bibinfo{pages}{103867}.
\bibitem[{Ma et~al.(2022)Ma, Zhou, Zhang, and Zhou}]{unfoldenoise}
\bibinfo{author}{C.~Ma}, \bibinfo{author}{J.~T. Zhou},
  \bibinfo{author}{X.~Zhang}, \bibinfo{author}{Y.~Zhou},
\newblock \bibinfo{title}{Deep unfolding for compressed sensing with denoiser},
\newblock in: \bibinfo{booktitle}{Int. Conf. Multimedia and Expo},
  \bibinfo{organization}{IEEE}, \bibinfo{year}{2022}, pp.
  \bibinfo{pages}{01--06}.
\bibitem[{Sun et~al.(2021)Sun, Dai, Li, Zou, and Xiong}]{csc}
\bibinfo{author}{J.~Sun}, \bibinfo{author}{W.~Dai}, \bibinfo{author}{C.~Li},
  \bibinfo{author}{J.~Zou}, \bibinfo{author}{H.~Xiong},
\newblock \bibinfo{title}{Compressive sensing via unfolded $l_0$-constrained
  convolutional sparse coding},
\newblock in: \bibinfo{booktitle}{Data Compress. Conf.},
  \bibinfo{organization}{IEEE}, \bibinfo{year}{2021}, pp.
  \bibinfo{pages}{183--192}.
\bibitem[{Behboodi et~al.(2022)Behboodi, Rauhut, and Schnoor}]{istagen}
\bibinfo{author}{A.~Behboodi}, \bibinfo{author}{H.~Rauhut},
  \bibinfo{author}{E.~Schnoor},
\newblock \bibinfo{title}{Compressive sensing and neural networks from a
  statistical learning perspective},
\newblock in: \bibinfo{booktitle}{Compressed Sensing in Information
  Processing}, \bibinfo{publisher}{Springer}, \bibinfo{year}{2022}, pp.
  \bibinfo{pages}{247--277}.
\bibitem[{Zhang and Ghanem(2018)}]{ista-net}
\bibinfo{author}{J.~Zhang}, \bibinfo{author}{B.~Ghanem},
\newblock \bibinfo{title}{{ISTA}-{N}et: Interpretable optimization-inspired
  deep network for image compressive sensing},
\newblock in: \bibinfo{booktitle}{Proc. Comput. Vis. and Pattern Recognit.},
  \bibinfo{organization}{IEEE}, \bibinfo{year}{2018}, pp.
  \bibinfo{pages}{1828--1837}.
\bibitem[{Sun et~al.(2016)Sun, Li, and Xu}]{admm-net}
\bibinfo{author}{J.~Sun}, \bibinfo{author}{H.~Li}, \bibinfo{author}{Z.~Xu},
\newblock \bibinfo{title}{Deep {ADMM}-{N}et for compressive sensing {MRI}},
\newblock \bibinfo{journal}{Advances Neural Inf. Process. Syst.}
  \bibinfo{volume}{29} (\bibinfo{year}{2016}).
\bibitem[{Ramirez et~al.(2021)Ramirez, Martinez-Torre, and Arguello}]{ladmm}
\bibinfo{author}{J.~M. Ramirez}, \bibinfo{author}{J.~I. Martinez-Torre},
  \bibinfo{author}{H.~Arguello},
\newblock \bibinfo{title}{{LADMM-N}et: an unrolled deep network for spectral
  image fusion from compressive data},
\newblock \bibinfo{journal}{Signal Process.} \bibinfo{volume}{189}
  (\bibinfo{year}{2021}) \bibinfo{pages}{108239}.
\bibitem[{Kouni et~al.(2022)Kouni, Paraskevopoulos, Rauhut, and
  Alexandropoulos}]{admmdad}
\bibinfo{author}{V.~Kouni}, \bibinfo{author}{G.~Paraskevopoulos},
  \bibinfo{author}{H.~Rauhut}, \bibinfo{author}{G.~C. Alexandropoulos},
\newblock \bibinfo{title}{{ADMM-DAD} net: a deep unfolding network for analysis
  compressed sensing},
\newblock in: \bibinfo{booktitle}{Int. Conf. Acoust., Speech and Signal
  Process.}, \bibinfo{organization}{IEEE}, \bibinfo{year}{2022}, pp.
  \bibinfo{pages}{1506--1510}.
\bibitem[{Ma et~al.(2019)Ma, Liu, Shou, and Yuan}]{tensoradmm}
\bibinfo{author}{J.~Ma}, \bibinfo{author}{X.-Y. Liu},
  \bibinfo{author}{Z.~Shou}, \bibinfo{author}{X.~Yuan},
\newblock \bibinfo{title}{Deep tensor admm-net for snapshot compressive
  imaging},
\newblock in: \bibinfo{booktitle}{Proc. IEEE/CVF Int. Conf. Comput. Vis.},
  \bibinfo{year}{2019}, pp. \bibinfo{pages}{10223--10232}.
\bibitem[{Zayyani et~al.(2015)Zayyani, Korki, and Marvasti}]{1bitcs}
\bibinfo{author}{H.~Zayyani}, \bibinfo{author}{M.~Korki},
  \bibinfo{author}{F.~Marvasti},
\newblock \bibinfo{title}{Dictionary learning for blind one bit compressed
  sensing},
\newblock \bibinfo{journal}{IEEE Signal Process. Lett.} \bibinfo{volume}{23}
  (\bibinfo{year}{2015}) \bibinfo{pages}{187--191}.
\bibitem[{Shen et~al.(2015)Shen, Li, Zhu, Cao, and Song}]{imagedl}
\bibinfo{author}{Y.~Shen}, \bibinfo{author}{J.~Li}, \bibinfo{author}{Z.~Zhu},
  \bibinfo{author}{W.~Cao}, \bibinfo{author}{Y.~Song},
\newblock \bibinfo{title}{Image reconstruction algorithm from compressed
  sensing measurements by dictionary learning},
\newblock \bibinfo{journal}{Neurocomputing} \bibinfo{volume}{151}
  (\bibinfo{year}{2015}) \bibinfo{pages}{1153--1162}.
\bibitem[{Li et~al.(2016)Li, Huang, and Misra}]{iot}
\bibinfo{author}{Z.~Li}, \bibinfo{author}{H.~Huang},
  \bibinfo{author}{S.~Misra},
\newblock \bibinfo{title}{Compressed sensing via dictionary learning and
  approximate message passing for multimedia internet of things},
\newblock \bibinfo{journal}{IEEE Internet of Things J.} \bibinfo{volume}{4}
  (\bibinfo{year}{2016}) \bibinfo{pages}{505--512}.
\bibitem[{Foucart and Rauhut(2013)}]{rf}
\bibinfo{author}{S.~Foucart}, \bibinfo{author}{H.~Rauhut},
\newblock \bibinfo{title}{An invitation to compressive sensing},
\newblock in: \bibinfo{booktitle}{A mathematical introduction to compressive
  sensing}, \bibinfo{publisher}{Springer}, \bibinfo{year}{2013}, pp.
  \bibinfo{pages}{1--39}.
\bibitem[{Candes et~al.(2011)Candes, Eldar, Needell, and Randall}]{cosparse}
\bibinfo{author}{E.~J. Candes}, \bibinfo{author}{Y.~C. Eldar},
  \bibinfo{author}{D.~Needell}, \bibinfo{author}{P.~Randall},
\newblock \bibinfo{title}{Compressed sensing with coherent and redundant
  dictionaries},
\newblock \bibinfo{journal}{Appl. and Comput. Harmon. Anal.}
  \bibinfo{volume}{31} (\bibinfo{year}{2011}) \bibinfo{pages}{59--73}.
\bibitem[{Cherkaoui et~al.(2018)}]{analvssyn}
\bibinfo{author}{H.~Cherkaoui}, et~al.,
\newblock \bibinfo{title}{Analysis vs synthesis-based regularization for
  combined compressed sensing and parallel mri reconstruction at 7 tesla},
\newblock in: \bibinfo{booktitle}{Eur. Signal Process. Conf.},
  \bibinfo{organization}{IEEE}, \bibinfo{year}{2018}, pp.
  \bibinfo{pages}{36--40}.
\bibitem[{Genzel et~al.(2021)Genzel, Kutyniok, and M{\"a}rz}]{genzel}
\bibinfo{author}{M.~Genzel}, \bibinfo{author}{G.~Kutyniok},
  \bibinfo{author}{M.~M{\"a}rz},
\newblock \bibinfo{title}{$l_1$-analysis minimization and generalized (co-)
  sparsity: When does recovery succeed?},
\newblock \bibinfo{journal}{Appl. and Comput. Harmon. Anal.}
  \bibinfo{volume}{52} (\bibinfo{year}{2021}) \bibinfo{pages}{82--140}.
\bibitem[{Shalev-Shwartz and Ben-David(2014)}]{shalev}
\bibinfo{author}{S.~Shalev-Shwartz}, \bibinfo{author}{S.~Ben-David},
  \bibinfo{title}{Understanding machine learning: From theory to algorithms},
  \bibinfo{publisher}{Cambridge university press}, \bibinfo{year}{2014}.
\bibitem[{Mohri et~al.(2018)Mohri, Rostamizadeh, and Talwalkar}]{mohri}
\bibinfo{author}{M.~Mohri}, \bibinfo{author}{A.~Rostamizadeh},
  \bibinfo{author}{A.~Talwalkar}, \bibinfo{title}{Foundations of machine
  learning}, \bibinfo{publisher}{MIT press}, \bibinfo{year}{2018}.
\bibitem[{Cao and Gu(2019)}]{gensgd}
\bibinfo{author}{Y.~Cao}, \bibinfo{author}{Q.~Gu},
\newblock \bibinfo{title}{Generalization bounds of stochastic gradient descent
  for wide and deep neural networks},
\newblock \bibinfo{journal}{Advances Neural Inf. Process. Syst.}
  \bibinfo{volume}{32} (\bibinfo{year}{2019}).
\bibitem[{Wang et~al.(2019)Wang, Diaz, Santos~Filho, and Calmon}]{wass}
\bibinfo{author}{H.~Wang}, \bibinfo{author}{M.~Diaz}, \bibinfo{author}{J.~C.~S.
  Santos~Filho}, \bibinfo{author}{F.~P. Calmon},
\newblock \bibinfo{title}{An information-theoretic view of generalization via
  wasserstein distance},
\newblock in: \bibinfo{booktitle}{Int. Symp. Inf. Theory (ISIT)},
  \bibinfo{organization}{IEEE}, \bibinfo{year}{2019}, pp.
  \bibinfo{pages}{577--581}.
\bibitem[{Arora et~al.(2018)Arora, Ge, Neyshabur, and Zhang}]{arora}
\bibinfo{author}{S.~Arora}, \bibinfo{author}{R.~Ge},
  \bibinfo{author}{B.~Neyshabur}, \bibinfo{author}{Y.~Zhang},
\newblock \bibinfo{title}{Stronger generalization bounds for deep nets via a
  compression approach},
\newblock in: \bibinfo{booktitle}{Int. Conf. Mach. Learn.},
  \bibinfo{organization}{PMLR}, \bibinfo{year}{2018}, pp.
  \bibinfo{pages}{254--263}.
\bibitem[{Bartlett et~al.(2017)Bartlett, Foster, and Telgarsky}]{spectralbound}
\bibinfo{author}{P.~L. Bartlett}, \bibinfo{author}{D.~J. Foster},
  \bibinfo{author}{M.~J. Telgarsky},
\newblock \bibinfo{title}{Spectrally-normalized margin bounds for neural
  networks},
\newblock \bibinfo{journal}{Advances Neur. Inf. Process. Syst.}
  \bibinfo{volume}{30} (\bibinfo{year}{2017}).
\bibitem[{Van~L. et~al.(2020)Van~L., Joukovsky, and Deligiannis}]{deeprnn}
\bibinfo{author}{H.~Van~L.}, \bibinfo{author}{B.~Joukovsky},
  \bibinfo{author}{N.~Deligiannis},
\newblock \bibinfo{title}{Interpretable deep recurrent neural networks via
  unfolding reweighted $\ell_1-\ell_1$ minimization: Architecture design and
  generalization analysis},
\newblock \bibinfo{journal}{arXiv preprint arXiv:2003.08334}
  (\bibinfo{year}{2020}).
\bibitem[{Joukovsky et~al.(2021)Joukovsky, Mukherjee, Van~L., and
  Deligiannis}]{unfoldrnn}
\bibinfo{author}{B.~Joukovsky}, \bibinfo{author}{T.~Mukherjee},
  \bibinfo{author}{H.~Van~L.}, \bibinfo{author}{N.~Deligiannis},
\newblock \bibinfo{title}{Generalization error bounds for deep unfolding rnns},
\newblock in: \bibinfo{booktitle}{Uncertain. Artif. Intell.},
  \bibinfo{organization}{PMLR}, \bibinfo{year}{2021}, pp.
  \bibinfo{pages}{1515--1524}.
\bibitem[{Vershynin(2018)}]{vershynin}
\bibinfo{author}{R.~Vershynin}, \bibinfo{title}{High-dimensional probability:
  An introduction with applications in data science},
  volume~\bibinfo{volume}{47}, \bibinfo{publisher}{Cambridge university press},
  \bibinfo{year}{2018}.
\bibitem[{Bartlett and Mendelson(2002)}]{bartlett}
\bibinfo{author}{P.~L. Bartlett}, \bibinfo{author}{S.~Mendelson},
\newblock \bibinfo{title}{Rademacher and gaussian complexities: Risk bounds and
  structural results},
\newblock \bibinfo{journal}{J. of Mach. Learn. Res.} \bibinfo{volume}{3}
  (\bibinfo{year}{2002}) \bibinfo{pages}{463--482}.
\bibitem[{LeCun et~al.(1998)LeCun, Bottou, Bengio, and Haffner}]{mnist}
\bibinfo{author}{Y.~LeCun}, \bibinfo{author}{L.~Bottou},
  \bibinfo{author}{Y.~Bengio}, \bibinfo{author}{P.~Haffner},
\newblock \bibinfo{title}{Gradient-based learning applied to document
  recognition},
\newblock \bibinfo{journal}{Proc. IEEE} \bibinfo{volume}{86}
  (\bibinfo{year}{1998}) \bibinfo{pages}{2278--2324}.
\bibitem[{Adcock et~al.(2016)Adcock, Hansen, and Roman}]{waveletcs}
\bibinfo{author}{B.~Adcock}, \bibinfo{author}{A.~C. Hansen},
  \bibinfo{author}{B.~Roman},
\newblock \bibinfo{title}{A note on compressed sensing of structured sparse
  wavelet coefficients from subsampled fourier measurements},
\newblock \bibinfo{journal}{IEEE Signal Process. Lett.} \bibinfo{volume}{23}
  (\bibinfo{year}{2016}) \bibinfo{pages}{732--736}.
\bibitem[{Gregor and LeCun(2010)}]{lista}
\bibinfo{author}{K.~Gregor}, \bibinfo{author}{Y.~LeCun},
\newblock \bibinfo{title}{Learning fast approximations of sparse coding},
\newblock in: \bibinfo{booktitle}{Proc. 27th Int. Conf. Mach. Learn.},
  \bibinfo{year}{2010}, pp. \bibinfo{pages}{399--406}.
\bibitem[{Li et~al.(2018)Li, Cheng, and Gui}]{robustadmmnet}
\bibinfo{author}{Y.~Li}, \bibinfo{author}{X.~Cheng}, \bibinfo{author}{G.~Gui},
\newblock \bibinfo{title}{Co-robust-{ADMM}-net: Joint {ADMM} framework and
  {DNN} for robust sparse composite regularization},
\newblock \bibinfo{journal}{IEEE Access} \bibinfo{volume}{6}
  (\bibinfo{year}{2018}) \bibinfo{pages}{47943--47952}.
\bibitem[{Kabanava et~al.(2015)Kabanava, Rauhut, and Zhang}]{robustanal}
\bibinfo{author}{M.~Kabanava}, \bibinfo{author}{H.~Rauhut},
  \bibinfo{author}{H.~Zhang},
\newblock \bibinfo{title}{Robust analysis $l_1$-recovery from gaussian
  measurements and total variation minimization},
\newblock \bibinfo{journal}{Eur. J. Appl. Math.} \bibinfo{volume}{26}
  (\bibinfo{year}{2015}) \bibinfo{pages}{917--929}.
\bibitem[{Kouni and Rauhut(2021)}]{star}
\bibinfo{author}{V.~Kouni}, \bibinfo{author}{H.~Rauhut},
\newblock \bibinfo{title}{Spark deficient gabor frame provides a novel analysis
  operator for compressed sensing},
\newblock in: \bibinfo{booktitle}{Int. Conf. Neural Inf. Process.},
  \bibinfo{organization}{Springer}, \bibinfo{year}{2021}, pp.
  \bibinfo{pages}{700--708}.
\bibitem[{Nam et~al.(2013)Nam, Davies, Elad, and Gribonval}]{cosparsity}
\bibinfo{author}{S.~Nam}, \bibinfo{author}{M.~E. Davies},
  \bibinfo{author}{M.~Elad}, \bibinfo{author}{R.~Gribonval},
\newblock \bibinfo{title}{The cosparse analysis model and algorithms},
\newblock \bibinfo{journal}{Appl. and Comput. Harmon. Anal.}
  \bibinfo{volume}{34} (\bibinfo{year}{2013}) \bibinfo{pages}{30--56}.
\bibitem[{Christensen(2003)}]{christensen}
\bibinfo{author}{O.~Christensen}, \bibinfo{title}{An introduction to frames and
  Riesz bases}, volume~\bibinfo{volume}{7}, \bibinfo{publisher}{Boston:
  Birkhäuser}, \bibinfo{year}{2003}.
\bibitem[{Casazza et~al.(2013)Casazza, Kutyniok, and Philipp}]{casazza}
\bibinfo{author}{P.~G. Casazza}, \bibinfo{author}{G.~Kutyniok},
  \bibinfo{author}{F.~Philipp},
\newblock \bibinfo{title}{Introduction to finite frame theory},
\newblock \bibinfo{journal}{Finite frames: theory and applications}
  (\bibinfo{year}{2013}) \bibinfo{pages}{1--53}.
\bibitem[{Zhu(2017)}]{genlasso}
\bibinfo{author}{Y.~Zhu},
\newblock \bibinfo{title}{An augmented {ADMM} algorithm with application to the
  generalized lasso problem},
\newblock \bibinfo{journal}{J. of Comput. and Graphical Statist.}
  \bibinfo{volume}{26} (\bibinfo{year}{2017}) \bibinfo{pages}{195--204}.
\bibitem[{Kabanava and Rauhut(2015)}]{kr}
\bibinfo{author}{M.~Kabanava}, \bibinfo{author}{H.~Rauhut},
\newblock \bibinfo{title}{Analysis $l_1$-recovery with frames and gaussian
  measurements},
\newblock \bibinfo{journal}{Acta Appl. Math.} \bibinfo{volume}{140}
  (\bibinfo{year}{2015}) \bibinfo{pages}{173--195}.
\bibitem[{Candes et~al.(2011)Candes, Eldar, Needell, and Randall}]{coh}
\bibinfo{author}{E.~J. Candes}, \bibinfo{author}{Y.~C. Eldar},
  \bibinfo{author}{D.~Needell}, \bibinfo{author}{P.~Randall},
\newblock \bibinfo{title}{Compressed sensing with coherent and redundant
  dictionaries},
\newblock \bibinfo{journal}{Appl. and Comput. Harmon. Anal.}
  \bibinfo{volume}{31} (\bibinfo{year}{2011}) \bibinfo{pages}{59--73}.
\bibitem[{Ledoux and Talagrand(2013)}]{tal}
\bibinfo{author}{M.~Ledoux}, \bibinfo{author}{M.~Talagrand},
  \bibinfo{title}{Probability in Banach Spaces: isoperimetry and processes},
  \bibinfo{publisher}{Springer Scie. \& Business Media}, \bibinfo{year}{2013}.
\bibitem[{Maurer(2016)}]{contraction}
\bibinfo{author}{A.~Maurer},
\newblock \bibinfo{title}{A vector-contraction inequality for rademacher
  complexities},
\newblock in: \bibinfo{booktitle}{Int. Conf. Algorithmic Learn. Theory},
  \bibinfo{organization}{Springer}, \bibinfo{year}{2016}, pp.
  \bibinfo{pages}{3--17}.
\bibitem[{Kouni and Panagakis(2023)}]{deconet}
\bibinfo{author}{V.~Kouni}, \bibinfo{author}{Y.~Panagakis},
\newblock \bibinfo{title}{Deconet: An unfolding network for analysis-based
  compressed sensing with generalization error bounds},
\newblock \bibinfo{journal}{IEEE Trans. Signal Process.} \bibinfo{volume}{71}
  (\bibinfo{year}{2023}) \bibinfo{pages}{1938--1951}.
\bibitem[{He et~al.(2015)He, Zhang, Ren, and Sun}]{he}
\bibinfo{author}{K.~He}, \bibinfo{author}{X.~Zhang}, \bibinfo{author}{S.~Ren},
  \bibinfo{author}{J.~Sun},
\newblock \bibinfo{title}{Delving deep into rectifiers: Surpassing human-level
  performance on imagenet classification},
\newblock in: \bibinfo{booktitle}{Proc. Int. Conf. Comput. Vis.},
  \bibinfo{organization}{IEEE}, \bibinfo{year}{2015}, pp.
  \bibinfo{pages}{1026--1034}.
\bibitem[{Ketkar(2017)}]{pytorch}
\bibinfo{author}{N.~Ketkar},
\newblock \bibinfo{title}{Introduction to pytorch},
\newblock in: \bibinfo{booktitle}{Deep learning with python},
  \bibinfo{publisher}{Springer}, \bibinfo{year}{2017}, pp.
  \bibinfo{pages}{195--208}.
\bibitem[{Kingma and Ba(2014)}]{adam}
\bibinfo{author}{D.~P. Kingma}, \bibinfo{author}{J.~Ba},
\newblock \bibinfo{title}{Adam: A method for stochastic optimization},
\newblock \bibinfo{journal}{arXiv preprint arXiv:1412.6980}
  (\bibinfo{year}{2014}).
\bibitem[{Prechelt(1998)}]{earlystop}
\bibinfo{author}{L.~Prechelt},
\newblock \bibinfo{title}{Early stopping-but when?},
\newblock in: \bibinfo{booktitle}{Neural Networks: Tricks of the trade},
  \bibinfo{publisher}{Springer}, \bibinfo{year}{1998}, pp.
  \bibinfo{pages}{55--69}.

\end{thebibliography}

\end{document}


\maketitle

\section*{Proof of Lemma~\ref{invab}}

\begin{proof}
    If $A+B$ is not invertible, then there exists some $x\neq0$ such that $Ax+Bx=0$. By assumption, $A$ is invertible, thus $-x=A^{-1}Bx$. Hence, $\|x\|=\|A^{-1}Bx\|\leq\|A^{-1}\|\|B\|\|x\|\overset{x\neq0}{\implies}1\leq\|A^{-1}\|\|B\|$, which contradicts our assumption, so $A+B$ is invertible. We also have: $A^{-1}(A+B)=I-(-A^{-1}B)\implies A+B=A(I+A^{-1}B)$. Since $A+B$ and $I+A^{-1}B$ are invertible, we get $(A+B)^{-1}=A^{-1}(I+A^{-1}B)^{-1}$. Due to the invertibility of $I+A^{-1}B$, we get
    \begin{align*}
        (I+A^{-1}B)^{-1}&+A^{-1}B(I+A^{-1}B)^{-1}=I\\
        \iff&(I+A^{-1}B)^{-1}=I-A^{-1}B(I+A^{-1}B)^{-1}\\
        \iff&\|(I+A^{-1}B)^{-1}\|=\|I-A^{-1}B(I+A^{-1}B)^{-1}\|\\
        \leq&\|I\|+\|A^{-1}B(I+A^{-1}B)^{-1}\|\\
        \leq&1+\|A^{-1}\|\|B\|\|(I+A^{-1}B)^{-1}\|\\
        \implies&\|(I+A^{-1}B)^{-1}\|\leq\frac{1}{1-\|A^{-1}\|\|B\|}.
    \end{align*}
    We apply the latter estimate to $\|(A+B)^{-1}\|\leq\|A^{-1}\|\|(I+A^{-1}B)^{-1}\|$ and the proof follows.
\end{proof}

\section*{Proof of Lemma~\ref{invsubtract}}

\begin{proof}
Since $B^{-1}-A^{-1}=B^{-1}(I-BA^{-1})=B^{-1}(AA^{-1}-BA^{-1})=B^{-1}(A-B)A^{-1}$, we deduce, by sub-multiplicativity of the norm $\|\cdot\|$, that $\|B^{-1}-A^{-1}\|\leq\|B^{-1}\|\|(A-B)\|\|A^{-1}\|$.
\end{proof}

\section*{Proof of Theorem~\ref{lipschitz}}

\begin{proof}
    In order to treat all layers in a uniform manner, we first set $f^0_{\Phi_1}(Y)=f^0_{\Phi_2}(Y)=Y$. For easiness in reading, we simply write $\Tilde{\Theta}_1$, $\Theta_1$, $\Lambda_1$, $W_1$, $B_1$ to denote the dependence on $\Phi_1$ (similarly for $\Phi_2$). Notice also that for two analysis operators $\Phi_1$, $\Phi_2$ associated to two different frames, we can always find a constant $0<\beta<\infty$ such that the frame operators of both frames are upper bounded by this $\beta$, i.e., $\|S_1\|_{\opnorm}\leq \beta $ and $\|S_2\|_{\opnorm}\leq \beta$, implying that $\|\Phi_1\|_{\opnorm}\leq\sqrt{\beta}$ and $\|\Phi_2\|_{\opnorm}\leq\sqrt{\beta}$, respectively. Hence, we can reasonably assume that both $\Phi_1$ and $\Phi_2$ lie in $\mathcal{F}_\beta$.\\
    Due to the explicit form of the matrices $\Tilde{\Theta}$, $\Theta$, $\Lambda$, $W$, $B$, the 1-Lipschitzness of $\st_{\lambda/\rho}(\cdot,\cdot)$ w.r.t. the first parameter and the introduction of mixed terms, we obtain
    \begin{align*}
    \|f^k_{\Phi_1}(Y)&-f^k_{\Phi_2}(Y)\|_F\\
    \leq&\|\Tilde{\Theta}_2f^{k-1}_{\Phi_2}(Y)+I'B_2+I''\st_{\lambda/\rho}(\Theta_2 f^{k-1}_{\Phi_2}(Y)+B_1)\\
    &-\Tilde{\Theta}_1f^{k-1}_{\Phi_1}(Y)-I'B_1-I''\st_{\lambda/\rho}(\Theta_1 f^{k-1}_{\Phi_1}(Y)+B)\|_F\\
    =&\|\Tilde{\Theta}_2f^{k-1}_{\Phi_2}(Y)-\Tilde{\Theta}_2f^{k-1}_{\Phi_1}(Y)+I'B_2+I''\st_{\lambda/\rho}(\Theta_2 f^{k-1}_{\Phi_2}(Y)+B_2)\\
    &-\Tilde{\Theta}_1f^{k-1}_{\Phi_1}(Y)+\Tilde{\Theta}_2f^{k-1}_{\Phi_1}(Y)-I'B_1-I''\st_{\lambda/\rho}(\Theta_1 f^{k-1}_{\Phi_1}(Y)+B_1)\|_F\\
    \leq&\|\Tilde{\Theta}_2-\Tilde{\Theta}_1\|_{\opnorm}\|f^{k-1}_{\Phi_1}(Y)\|_F+\|\Tilde{\Theta}_2\|_{\opnorm}\|f^{k-1}_{\Phi_2}(Y)-f^{k-1}_{\Phi_1}(Y)\|_F+\|I'\|_{\opnorm}\\
    &\cdot\big(\|B_2-B_1\|_F+\|\Theta_2f^{k-1}_{\Phi_2}(Y)-\Theta_2f^{k-1}_{\Phi_1}(Y)\\
    &+\Theta_2f^{k-1}_{\Phi_1}(Y)-\Theta_1f^{k-1}_{\Phi_1}(Y)\|_F+\|B_2-B_1\|_F\big)\\
    \leq&\|\Lambda_2-\Lambda_1\|_{\opnorm}\|f^{k-1}_{\Phi_1}(Y)\|_F+\|\Lambda_2\|_{\opnorm}\|f^{k-1}_{\Phi_2}(Y)-f^{k-1}_{\Phi_1}(Y)\|_F\\
    &+2(\|\Theta_2\|_{\opnorm}\|f^{k-1}_{\Phi_2}(Y)-f^{k-1}_{\Phi_1}(Y)\|_F+\|\Theta_2-\Theta_1\|_{\opnorm}\|f^{k-1}_{\Phi_1}(Y)\|_F)\\
    &+3\|B_2-B_1\|_F\\
    \leq&3(1+2\|W_2\|_{\opnorm})\|f^{k-1}_{\Phi_2}(Y)-f^{k-1}_{\Phi_1}(Y)\|_F+3\|B_2-B_1\|_F\\
    &+6\|W_2-W_1\|_{\opnorm}\|f^{k-1}_{\Phi_1}(Y)\|_F\\
    \leq&3(1+2\beta q\rho)\|f^{k-1}_{\Phi_2}(Y)-f^{k-1}_{\Phi_1}(Y)\|_F+3\underbrace{\|B_2-B_1\|_F}_{\heartsuit}\\
    &+6\underbrace{\|W_2-W_1\|_{\opnorm}}_{(\ast)}\|f^{k-1}_{\Phi_1}(Y)\|_F.
\end{align*}
\noindent We treat ($\ast$) and ($\heartsuit$) separately. We start with the former:
\begin{align*}
    \|W_2-W_1\|_{\opnorm}\leq\|&\rho\Phi_2(A^TA+\rho \Phi^T_2\Phi_2)^{-1}\Phi^T_2-\rho\Phi_1(A^TA+\rho \Phi^T_1\Phi_1)^{-1}\Phi^T_1\|_{\opnorm}\\
    =\rho&\|\Phi_2(A^TA+\rho \Phi^T_2\Phi_2)^{-1}\Phi^T_2-\Phi_2(A^TA+\rho \Phi^T_1\Phi_1)^{-1}\Phi^T_2\\
    &+ \Phi_2(A^TA+\rho \Phi^T_1\Phi_1)^{-1}\Phi^T_2-\Phi_1(A^TA+\rho \Phi^T_1\Phi_1)^{-1}\Phi^T_1\|_{\opnorm}\\
    \leq\rho&\underbrace{\|\Phi_2[(A^TA+\rho \Phi^T_2\Phi_2)^{-1}-(A^TA+\rho \Phi^T_1\Phi_1)^{-1}]\Phi^T_2\|_{\opnorm}}_{(\star)}\\
    +&\rho\underbrace{\|\Phi_2(A^TA+\rho \Phi^T_1\Phi_1)^{-1}\Phi^T_2-\Phi_1(A^TA+\rho \Phi^T_1\Phi_1)^{-1}\Phi^T_1\|_{\opnorm}}_{(\dagger)}.
\end{align*}
According to the proof of Proposition~\ref{boundedoutput}, we have
$\|(A^TA+\rho \Phi^T_1\Phi_1)^{-1}\|_{\opnorm}=\|(A^TA+\rho \Phi^T_2\Phi_2)^{-1}\|_{\opnorm}\leq q$, with $q$ defined as in the latter Proposition.\\
\noindent For the term $(\dagger)$ we obtain
\begin{align*}
    \|\Phi_2&(A^TA+\rho \Phi^T_1\Phi_1)^{-1}\Phi^T_2-\Phi_1(A^TA+\rho \Phi^T_1\Phi_1)^{-1}\Phi^T_1\|_{\opnorm}\\
    =&\|\Phi_2(A^TA+\rho \Phi^T_1\Phi_1)^{-1}\Phi^T_2-\Phi_2(A^TA+\rho \Phi^T_1\Phi_1)^{-1}\Phi^T_1\\
    &+\Phi_2(A^TA+\rho \Phi^T_1\Phi_1)^{-1}\Phi^T_1-\Phi_1(A^TA+\rho \Phi^T_1\Phi_1)^{-1}\Phi^T_1\|_{\opnorm}\\
    \leq&\|\Phi_2\|_{\opnorm}\|(A^TA+\rho \Phi^T_1\Phi_1)^{-1}\|_{\opnorm}\|\Phi_2-\Phi_1\|_{\opnorm}\\
    &+\|\Phi_1\|_{\opnorm}\|(A^TA+\rho \Phi^T_1\Phi_1)^{-1}\|_{\opnorm}\|\Phi_2-\Phi_1\|_{\opnorm}\\
    \leq&2q\sqrt{\beta }\|\Phi_2-\Phi_1\|_{\opnorm},
\end{align*}
Regarding $(\star)$, we have
\begin{align*}
    \|\Phi_2&[(A^TA+\rho \Phi^T_2\Phi_2)^{-1}-(A^TA+\rho \Phi^T_1\Phi_1)^{-1}]\Phi^T_2\|_{\opnorm}\\
    \leq&\beta \|(A^TA+\rho \Phi^T_2\Phi_2)^{-1}-(A^TA+\rho \Phi^T_1\Phi_1)^{-1}\|_{\opnorm}\\
    \leq&\beta \rho\|(A^TA+\rho\Phi_1^T\Phi_1)^{-1}\|_{\opnorm}\|(A^TA+\rho\Phi_2^T\Phi_2)^{-1}\|_{\opnorm}\|\Phi_2^T\Phi_2-\Phi_1^T\Phi_1\|_{\opnorm}\\
    \leq&2\beta ^{3/2}q^2\rho\|\Phi_2-\Phi_1\|_{\opnorm}.
\end{align*}
\noindent Hence, for the term $\|W_2-W_1\|_{\opnorm}$ we get
\begin{align*}
\|W_2-W_1\|_{\opnorm}&\leq(2\beta \sqrt{\beta }q^2\rho^2+2q\rho\sqrt{\beta })
\|\Phi_2-\Phi_1\|_{\opnorm}\notag\\
&=2q\rho\sqrt{\beta }(1+\beta q\rho)\|\Phi_2-\Phi_1\|_{\opnorm}. 
\end{align*}
We proceed with estimating ($\heartsuit$):
\begin{align*}
    \|B_2-&B_1\|_F=\|\Phi_2(A^TA+\rho \Phi^T_2\Phi_2)^{-1}A^Ty-\Phi_1(A^TA+\rho\Phi^T_1\Phi_1)^{-1}A^Ty\|_{\opnorm}\\
    \leq&\|A\|_{\opnorm}\|Y\|_F\|\Phi_2(A^TA+\rho\Phi^T_2\Phi_2)^{-1}-\Phi_1(A^TA+\rho\Phi^T_1\Phi_1)^{-1}\|_{\opnorm}\\
    \leq&\|A\|_{\opnorm}\|Y\|_F\|\Phi_2(A^TA+\rho\Phi^T_2\Phi_2)^{-1}-\Phi_2(A^TA+\rho\Phi^T_1\Phi_1)^{-1}\\
    &+\Phi_2(A^TA+\rho\Phi^T_1\Phi_1)^{-1}-\Phi_1(A^TA+\rho\Phi^T_1\Phi_1)^{-1}\|_{\opnorm}\\
    \leq&\|A\|_{\opnorm}\|Y\|_F\bigg(\sqrt{\beta }\|(A^TA+\rho\Phi^T_2\Phi_2)^{-1}\\
    &-(A^TA+\rho\Phi^T_1\Phi_1)^{-1}\|_{\opnorm}+q\|\Phi_2-\Phi_1\|_{\opnorm}\bigg)\\
    &\underset{Lemma~\ref{invsubtract}}{\leq}\|A\|_{\opnorm}\|Y\|_F(2q^2\beta\rho+q)\|\Phi_2-\Phi_1\|_{\opnorm}\\
    =&q(1+2q\beta\rho)\|A\|_{\opnorm}\|Y\|_F\|\Phi_2-\Phi_1\|_{\opnorm}.
\end{align*}
\noindent Putting everything together yields
\begin{multline}
\label{gd}
    \|f^k_{\Phi_1}(Y)-f^k_{\Phi_2}(Y)\|_F\leq3(1+2\beta q\rho)\|f^{k-1}_{\Phi_1}(Y)-f^{k-1}_{\Phi_2}(Y)\|_F\\
    +\bigg(12q\rho\sqrt{\beta }(1+\beta q\rho)\|f^{k-1}_{\Phi_1}(Y)\|_F+3q(1+2q\beta \rho)\|A\|_{\opnorm}\|Y\|_F\bigg)\|\Phi_2-\Phi_1\|_{\opnorm}\\
    \underset{Prop.~\ref{boundedoutput}}{\leq}G\|f^{k-1}_{\Phi_1}(Y)-f^{k-1}_{\Phi_2}(Y)\|_F+E_k\|\Phi_2-\Phi_1\|_{\opnorm},
\end{multline}
where
\begin{align}
G&=3(1+2\beta q\rho),\\
D_0&=0,\quad D_k=\sum_{i=0}^{k-1}G^i,\quad k\geq1,\\
E_k&=\|A\|_{\opnorm}\|Y\|_F\bigg(qG+36q^2\rho \beta (1+\beta q\rho)D_{k-1}\bigg),\quad k\geq1.
\end{align}

We employ the previous definitions of $G, E_k$ and prove via induction that
\begin{equation}\label{kl}
    K_L=\sum_{k=1}^LG^{L-k}E_k,\quad L\geq1.
\end{equation}
First, notice that for $L=1$ it holds
\begin{align*}
    \|f^1_{\Phi_2}(Y)-f^1_{\Phi_1}(Y)\|_F&=\|I'B_1+I''\st_{\lambda/\rho}(B_1)-I'B_2-I''\st_{\lambda/\rho}(B_2)\|_F\\
    &\leq3\|B_2-B_1\|_F\\
    &\leq3q(1+2q\beta \rho)\|A\|_{\opnorm}\|Y\|_F\|\Phi_2-\Phi_1\|_{\opnorm}\\
    &=E_1\|\Phi_2-\Phi_1\|_{\opnorm}\\
    &=K_1\|\Phi_2-\Phi_1\|_{\opnorm},
\end{align*}
so that $K_1$ has indeed the form described in \eqref{kl}. Suppose that \eqref{kl} holds for some $L\in\mathbb{N}$. Then, for $L+1$:
\begin{align*}
    \|f^{L+1}_{\Phi_2}(Y)-f^{L+1}_{\Phi_1}(Y)\|_F&\leq G\|f^{L}_{\Phi_2}(Y)-f^{L}_{\Phi_1}(Y)\|_F+E_{L+1}\|\Phi_2-\Phi_1\|_{\opnorm}\\
    &\leq(GK_L+E_{L+1})\|\Phi_2-\Phi_1\|_{\opnorm}\\
    &=\left(G\sum_{k=1}^LG^{L-k}E_k+E_{L+1}\right)\|\Phi_2-\Phi_1\|_{\opnorm}\\
    &=\left(\sum_{k=1}^{L+1}G^{L-k}E_k\right)\|\Phi_2-\Phi_1\|_{\opnorm}\\
    &=K_{L+1}\|\Phi_2-\Phi_1\|_{\opnorm}.
\end{align*}
The proof follows.
\end{proof}